\documentclass[10pt, conference, compsocconf]{IEEEtran}
\usepackage{mathrsfs}
\usepackage[noadjust]{cite}
\IEEEoverridecommandlockouts
\usepackage{epsfig}
\usepackage{amssymb,latexsym,amsfonts,amsmath}
\usepackage{graphicx}
\usepackage{subfigure}
\usepackage{color}
\usepackage{url}
\usepackage{amsmath}
\usepackage{amsthm}
\usepackage{amssymb}
\usepackage{epstopdf}
\begin{document}
	\title{Encoding Multi-Resolution Brain Networks Using Unsupervised Deep Learning}
    \author{\IEEEauthorblockN{Arash Rahnama\IEEEauthorrefmark{1}, Abdullah Alchihabi\IEEEauthorrefmark{2}, Vijay Gupta\IEEEauthorrefmark{1}, Panos J. Antsaklis,\IEEEauthorrefmark{1}} Fatos T. Yarman Vural\IEEEauthorrefmark{2}
	\IEEEauthorblockA{\IEEEauthorrefmark{1} Department of Electrical Engineering\\
	University of Notre Dame, Notre Dame, IN 46556, USA\\ Email:~\{arahnama,~vgupta2,~antsaklis.1\} @nd.edu.}
	\IEEEauthorblockA{\IEEEauthorrefmark{2} Department of Computer Engineering, Middle East Technical University, 06800 Ankara, Turkey\\
	Email: {abdullah.alchihabi@metu.edu.tr},~{vural@ceng.metu.edu.tr}}}
	\maketitle
\begin{abstract}
The main goal of this study is to extract a set of brain networks in multiple time-resolutions to analyze the connectivity patterns among the anatomic regions for a given cognitive task. We suggest a deep architecture which learns the natural groupings of the connectivity patterns of human brain in multiple time-resolutions. The suggested architecture is tested on task data set of Human Connectome Project (HCP) where we extract multi-resolution networks, each of which corresponds to a cognitive task. At the first level of this architecture, we decompose the fMRI signal into multiple sub-bands using wavelet decompositions. At the second level, for each sub-band, we estimate a brain network extracted from short time windows of the fMRI signal. At the third level, we feed the adjacency matrices of each mesh network at each time-resolution into an unsupervised deep learning algorithm, namely, a Stacked De-noising Auto-Encoder (SDAE). The outputs of the SDAE provide a compact connectivity representation for each time window at each sub-band of the fMRI signal. We concatenate the learned representations of all sub-bands at each window and cluster them by a hierarchical algorithm to find the natural groupings among the windows. We observe that each cluster represents a cognitive task with a performance of $93\%$ Rand Index and $71\%$ Adjusted Rand Index. We visualize the mean values and the precisions of the networks at each component of the cluster mixture. The mean brain networks at cluster centers show the variations among cognitive tasks and the precision of each cluster shows the within cluster variability of networks, across the subjects.
\end{abstract}
\section{Introduction} \label{sec:intro}
The data produced by functional Magnetic Resonance Imaging (fMRI) is high-dimensional and sometimes not suitable for analyzing the cognitive states \cite{firat2015learning}. Learning efficient low-dimensional features from high-dimensional complex input spaces is crucial for the decoding of cognitive processes. In this paper, we explore deep learning algorithms in order to i) find a compact  representation of connectivity patterns embedded in fMRI signals, ii) detect natural groupings of these patterns and, iii) use these natural groups to extract brain networks to represent cognitive tasks.

Our framework is built upon our previous work in the area \cite{onal2016hierarchical}, where we decompose fMRI signals into various frequency sub-bands using their wavelet transforms. We further utilize the signals at different sub-bands to form multi-resolution brain networks. Recent studies have shown that brain networks formed by the correlation of voxel pairs' in fMRI signals provide more information for brain decoding compared to the temporal information of single voxels \cite{lindquist2008statistical,richiardi2013machine}. Moreover, there has been a shift in the literature toward brain decoding algorithms that are based on the connectivity patterns in the brain motivated by the belief that these patterns provide more information about cognitive tasks than the isolated behavior of individual anatomic regions \cite{shirer2012decoding,ekman2012predicting,onal2017new}. 

Contrary to the methods suggested in \cite{lindquist2008statistical,richiardi2013machine} where supervised learning algorithms are employed for brain decoding, in this paper, we investigate the common groupings in HCP task data set to find out if these natural groups correspond to the cognitive tasks. This approach enables us to find shared network representations of a cognitive task together with its variations across the subjects. Additionally, multi-resolution representation of the fMRI signals enables us to observe the variations of networks among different frequency sub-bands. 

After constructing the brain networks representing the connectivity patterns among the anatomic regions of the brain at each sub-level, a Stacked De-noising Auto-Encoder (SDAE) algorithm is employed to learn shared connectivity features associated with a task based on the estimated mesh networks at different sub-bands. We concatenate the learned connectivity patterns from several wavelet sub-bands and utilize them in a hierarchical clustering algorithm with a distance matrix based on their correlations. The main reason behind concatenation of the feature matrices is that the detected patterns in the brain at different frequencies provide complementary information in regard to the overall cognitive state of the brain. 

Our results show that the mesh network representation of cognitive tasks is superior compared to fMRI time-series representation. We observe that SDAE successfully learns a set of connectivity patterns which provide an increased clustering performance. The performances are further improved by fusing the learned representations from multiple time-resolutions. This shows that the modeling of the connectivity of brain in multiple sub-bands of the fMRI signal leads to diverse mesh networks carrying complementary information for representing the cognitive tasks. The high rand index $93\%$ obtained at the output of the clustering algorithm proves the existence of natural groups with low within-cluster-variances and high between-class-variances among the tasks. 

In order to analyze the similarities and distinctions among the network topologies of fMRI signal, we visualize the networks and their precisions at the cluster centers. The cluster precisions, indicate shared connectivities among the subjects, whereas the mesh networks at the cluster center show a representative network for each cognitive task. It is observed that there are high inter-subject variances in the mesh networks.
\section{Experimental Setup} \label{sec:exp}
We use the fMRI data from HCP for $300$ subjects performing specific tasks. A subject performs seven distinct cognitive tasks during the experiment given in Table \ref{tab:firsttable} \cite{barch2013function}. Each task $t$ consists of $s_t$ scans of the brain volume representing the changes in the brain during the task (the underlying cognitive process). The duration and the number of scans are task-dependent but the same for all participants. The total number of scans is $S=1940$ and we use $R = 90$ anatomical regions of $116$ AAL after removing the anatomical regions in Cerebellum and Vermis. 
\begin{table}
	\centering
	\resizebox{.80\textwidth}{!}{\begin{minipage}{\textwidth}
			\begin{tabular}{l*{7}{c}r}
				\hline
				& Emotion & Gambling & Language & Motor & Relational & Social &  WM\\
				\hline
				Scans & 176 & 253 & 316 & 284  & 232 & 274 & 405\\
				\hline
				Durations & 2:16 & 3:12 & 3:57 & 3:34& 2:56 & 3:27 & 5:01\\
			\end{tabular}
	\end{minipage}}
	\caption{Scans per Task and the Duration for each Task (min:sec).}
	\label{tab:firsttable}  
\end{table}
Representative time-series data points are attained by spatially averaging the signals associated with voxels ($n$) residing in the same region ($r$) in the brain,
\begin{align*}
X_r(t)=\frac{1}{N} \sum_{\forall n \in r}^{} X_n(t),
\end{align*}
where $N$ represents the total number of voxels in region $r$.
\section{Hierarchical Multi-Resolution Mesh Networks (HMMNs)} \label{sec:hmmns}
\begin{figure*}[!t]
	\centering
	\includegraphics[scale = 0.5]{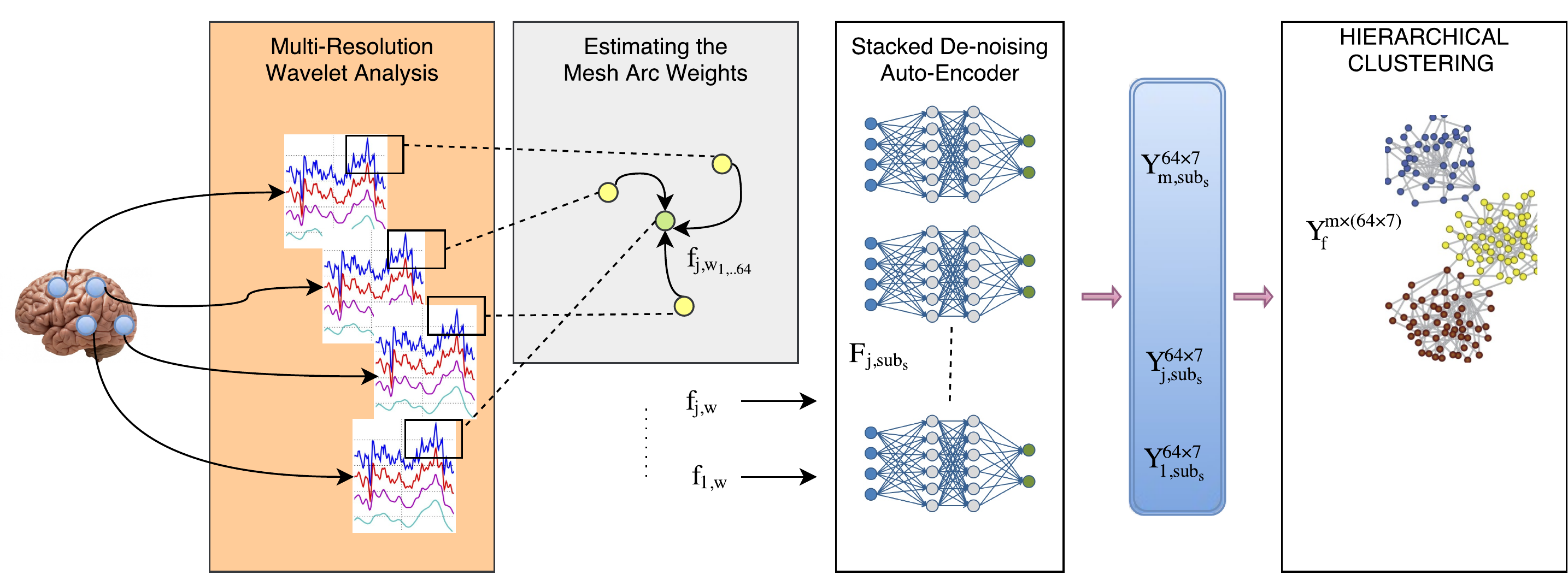}
	\caption{An Overview of the Proposed Deep Learning Framework.}
	\label{fig:Diagram}
\end{figure*}
In our work, we utilize the representative time series obtained for each anatomic region to build a set of local meshes. The local meshes estimated around each anatomic region are ensembled to form a mesh network. This is motivated by the fact that the structure of the brain is highly interconnected and that neurons influence each other based on the strengths of their synaptic connections \cite{pantazatos2012decoding}. HMMNs model cognitive tasks by estimating a mesh network at each frequency sub-band. It is expected that the brain network at each sub-band provides supplementary information about the underlying brain activities. We will show that our modeling of the brain with HMMNs greatly enhances the brain decoding performance by allowing us to look at cognitive states of the brain regions in multiple time-resolutions.

As the first step, the representative time-series $X_r(t)$, for each anatomic region $r$ are decomposed into a set of signals in different time-resolutions. This allows us to estimate and analyze how the anatomical regions process information in different frequency resolutions \cite{thompson2015frequency}. We adopt Discrete Wavelet Transform (DWT) as our main tool \cite{bullmore2004wavelets}. We apply the DWT to $X_r(t)$ for all brain regions to decompose the signals into $l$ sub-bands where $l=1,2,...,11$ ($L=11$). At sub-band level $l$, we attain two sets of orthonormal components named as sets of approximation coefficients $\mathcal{A}=\{a_{r,l,k}\}$ and detail coefficients $\mathcal{D}=\{d_{r,l,k}\}$ where $k$ represents the location of the wavelet waveform in discrete-time \cite{bullmore2004wavelets}. These coefficients then may be utilized to reconstruct the fMRI signals at each frequency level, yielding the total of $(2\times L)+1$ fMRI time-series. Formally, the representative time-series at sub-band $j$ ($j\in [0,1,...,2L]$) may be defined as,
\begin{align*}
x_{j,r}(t)=
\begin{cases}
X_r(t), & \text{if}\ j=0 \\
\sum_{k}^{}a_{r,l,k}\Phi_{l,k}(t)~and~ l=j & \text{if}\ 1\leq j\leq L\\
\sum_{k}^{}d_{r,l,k}\Psi_{l,k}(t)~and~ l=j-L+1 & \text{if}\ j > L
\end{cases}
\end{align*}
where $\Phi_{l,k}$ and $\Psi_{l,k}$ are called the mother wavelet and the father wavelet. More details on our approach are given in \cite{onal2016hierarchical}. 

Now, we can construct a mesh network at each sub-band to represent cognitive tasks in terms of the relationships among anatomic regions. The construction of these networks help us analyze the topological properties of the brain and extract connectivity patterns associated with a cognitive process at each sub-band. In order to demonstrate the benefits of our approach, we propose an unsupervised clustering framework which can successfully take advantage of the connectivity patterns and distinguish between different cognitive tasks at multiple sub-bands. For this purpose, we divide the entire experiment session ($S=1940$ number of scans) for a subject into unlabeled windows of length $w_i=30$ consisting of $30$ discrete scans, where $i=1,...,64$ for each subject for the entire experiment. The length of the window is determined empirically, as the shortest time-interval which provides the highest rand index, at the output of clustering. It is important to note that the windows are unlabeled and may consist of overlapping data points from different cognitive tasks. 

The nodes of the mesh networks are connected to their $p$-nearest neighbors to form a star mesh around a region. The nearest $p$ neighbors for a certain node are the ones having the largest Pearson correlation coefficients with the node. For each mesh formed around an anatomic region $r$, the arc weights for the window $w_i$ are estimated at the sub-band $j$ using the following regularized linear model,
\begin{align} \label{mesh}
x_{j,w_i,r}=\sum_{r^\prime \in N_p[r]}^{} a_{j,w_i,r,r^\prime} x_{j,w_i,r^\prime} + \lambda |a_{j,w_i,r,r^\prime}|^2 + \epsilon_{j,w_i,r}
\end{align}
where the regularization parameter is $\lambda$. The mesh arc weights $a_{j,w_i,r,r^\prime}$, defined in the $N_p[r]$ neighborhood of region $r$, are estimated by minimizing the error $\epsilon_{j,w_i,r}$. $x_{j,w_i,r}$ is a vector representing the average voxel time-series in region $r$ at sub-band $j$ for the window $w_i$, such that,
\begin{align*}
x_{j,w_i,r}=[x_{j,w_i,r}(1),x_{j,w_i,r}(2),...,x_{j,w_i,r}(30)].
\end{align*}

The relation defined in (\ref{mesh}) is solved for each region $r$ with its neighbors separately. In other words, we obtain an independent local mesh around each region $r$. After estimating all the mesh arc weights, we put them under the vector $A_{j,w_i} = \{a_{j,w_i,r,r^\prime}\}_{r,r^\prime}^R$, called Mesh Arc Descriptors (MADs). We represent $G_{j,w_i}$ as an ensemble of all local meshes. Lastly, the mesh networks are estimated for the original fMRI signal, and its approximation and detail parts of different resolutions. Consequently, we form $2L + 1$ distinct mesh networks for the frequency sub-bands $\{A_0,A_1,A_2,...,A_L,D_1,D_2,...,D_L\}$.  

The multi-resolution mesh network for a subject is defined by a connectivity graph, $G_{j,w_i} = \{ V, A_{j,w_i}: \forall j \}$, for each unlabeled window $w_i$ and for each sub-band $j$. The set of vertices $V$ corresponds to the ordered set of anatomic regions and is of size $R$. Vertex attributes are the time-series $x_{j,w_i,r}$ contained in the window $w_i$, at the sub-band $j$. The arc weights, $A_{j,w_i} = \{a_{j,w_i,r,r^\prime}\}_{r,r^\prime}^R$ between regions $r$ and $r^\prime$, for each window $w_i$ are obtained from the local meshes of the representative time-series data points at sub-band $j$. This process results in $2L + 1$ distinct mesh networks represented by an adjacency matrix of size $R\times R$ made up of ($\forall_{r,r^\prime} a_{j,w_i,r,r^\prime}$) for each window $w_i$ ($i=1,...,64$). We concatenate the arc weights under a vector ($f_{j,w_i}$) of size $1 \times R^2$ and embed the brain network for the window $w_i$ at sub-band $j$. This means that for each level $j$ and each subject, we represent the entire experiment by a large unlabeled matrix of size $64 \times (R^2)$ i.e. $F_{j,sub_s}=[f_{j,w_1},...,f_{j,w_{64}}]^T$. Next, we will introduce a deep learning algorithm which learns a set of compact connectivity patters from the embedded brain networks and consequently, cluster the windows of similar connectivity patterns. Each cluster of similar connectivity patterns represent a specific cognitive task (see Table \ref{tab:firsttable}).
\section{The Deep Learning Architecture} \label{sec:dl}
The embedded mesh networks model the connectivity among the anatomic regions at different sub-bands of fMRI signal under each window $w_i$ for each subject. Next, we utilize a deep learning architecture to extract a set of compact connectivity patterns from the mesh networks. We will show that the learned connectivity patterns form natural clusters corresponding to cognitive states. To meet this goal, we design a multi-layer stacked de-noising sparse auto-encoder \cite{poultney2007efficient}.
For each sub-band $j$, we train a SDAE that takes the windows in the embedded brain network associated with subject $sub_s$ i.e. $f_{j,w_i} \in F_{j,sub_s}, i=1,...,64$ as its input, and produces a vector $y$ of size $1\times 7$. Recall that there are a total of $7$ cognitive tasks. The learned features represent the connectivity patterns at sub-band $j$ for subject $sub_s$ as follows,
\begin{align*}
Y_\theta(F_{j,sub_s})=S(W F_{j,sub_s}+B),
\end{align*} 
 with the auto-encoder parameter set $\theta=[W,B]$ where $W$ is the collection of weights $\{W_i\}_{1:4}$, $B$ is the collection of biases $\{B_i\}_{1:4}$ at each neuron and $S$ represents the activation function $\arctan$. Our sparse auto-encoder design includes an input layer of size $f_{j,w_i}^T$ with three hidden layers $[500,64,21]$ and an output layer of size $7$ and the sparsity parameter $\rho$. The output of each neuron $y_i$ may be represented as $y_i=\sum_{j=1}^{n} w_jx_j + b_i$, where $n$ and $x_j$'s indicate the total number of neurons and the neurons' outputs from the previous layer. The objective function $J$ is to minimize the mean-squared loss function $L(W,B|F_{j,sub_s})$ in the presence of an $L_2$-Ridge regularization with parameter $\lambda_2$ which adds stability and robustness to the learning process,
\begin{align*}
J=\arg \min_{w_i,b_i}\{ L(W,B|F_{j,sub_s}) + \lambda_2 ||W||_2^2 \}.
\end{align*}

In order to deal with the possible noise in the input data points, we follow a dropout training procedure based on which at each learning epoch, $20\%$ of the data points are removed. It has been shown that this de-noising procedure will control for over-fitting, as reported in \cite{wager2013dropout}. After training the above auto-encoder, one can extract the feature matrices for subject $sub_s$ at sub-band $j$ to attain $({Y_{j,sub_s}^{(64 \times 7)}})_f$. Our results will show that our proposed deep learning algorithm is capable of removing the large intra-variance amongst input data points and can give an effective representation of the brain networks in a low-dimensional space. This can be considered as a non-linear mapping model from a high-dimensional space to a low-dimensional space suited for clustering.
\section{Hierarchical Clustering}
The main objective behind our work is to design a data driven cluster analysis that is suitable for discriminating between distinct connectivity patterns associated with given cognitive states at different frequency levels. We perform a hierarchical clustering on a combination of features from different frequency levels attained from the deep learning algorithm to show that the framework is capable of detecting the cognitive tasks (given in Table \ref{tab:firsttable}) based on their connectivity manifestation in the brain networks and their learned features after the deep learning architecture.

The clustering algorithm clusters a subject's brain features matrix $Y_f^{64 \times (m \times 7)}=[({Y_{1,sub_s}^{(64 \times 7)}})_f,...,({Y_{m,sub_s}^{(64 \times 7)}})_f]$ consisting of the concatenation of the feature matrices from $m$ different frequency levels selected from the the frequency sub-bands $\{A_0,A_1,A_2,...,A_{11},D_1,D_2,...,D_{11}\}$. This is to show that each frequency level carries complementary information in regard to cognitive tasks performed during the experiment. Given the $w=64$ discrete-time windows, the clustering algorithm attempts to divide the data points into $k=7$ clusters ($c_k$, $k=1,...,7$), by minimizing the following cost function,
\begin{align*}
V=\sum_{k=1}^{7} V_k = \sum_{k=1}^{7}(\sum_{y_j \in c_k \cap y_j\in Y_f}^{}dis(y_j,c_k)),
\end{align*}
where the distance matrix $dis(y_j,c_k)$ is based on the Pearson Correlation between data points which closely models the functional connectivity pattern in the brain from one task to another. The exact relation between the distance matrix and the correlation matrix is,
\begin{align*}
dis(y_j,c_k)=1-Corr^2(y_j,y_k).
\end{align*}

The entries of the correlation matrix $Corr(y_j,y_k)$ indicate the degree to which window $y_j$ is correlated with window $y_k$. The above relation can capture the time-varying coupling between windows and consequently closely model the flow of change in brain features from one cognitive state to another \cite{calhoun2014chronnectome}. Fig. \ref{fig:Diagram} depicts the entire deep learning framework. After clustering the unlabeled windows into $7$ different clusters, in the next section, we will compare the resulting clusters with the labeled data points given in Table \ref{tab:firsttable} in order to examine the performance of our proposed approach. We utilize Rand Index (RI) and Adjusted Rand Index (ARI) as performance measures for our algorithm \cite{milligan1986study}.
\section{Experimental Results} \label{sec:expres}
In this section, we test the validity of the suggested deep learning architecture in two groups of experiments. The first set of experiments measures the cluster validity by clustering the fMRI signal, mesh arc weights of single and multi-resolution signals and utilizing the measures of performance RI and ARI. The second group of experiments visualizes the mesh networks obtained across subjects and cognitive tasks to observe the inter-task and inter-subject variabilities. We perform within-subject clustering analyses based on the fMRI signals collected from $100$ subjects (described in Section \ref{sec:exp}). The design parameters are selected empirically through a cross validation process based on performance. We search for the optimal design parameters based on the sets, $p \in \{10,20,30,40,50\}$, $\lambda \in \{16,32,128,256\}$, $\rho \in \{0.01,0.001,0.0001\}$, and $\lambda_2 \in \{0.00001,0.00055,0.0001\}$. We select the design parameters, $p=40$ and $\lambda=32$ for the mesh networks (Section \ref{sec:hmmns}), $\rho=0.001$ and $\lambda_2=0.00055$ for the SDAE design (Section \ref{sec:dl}) as optimal values. The RI and ARI values given in the tables for each experiment describe the average clustering performance for all $100$ subjects. 

Table \ref{tab:table4} gives a performance comparison between the clustering of the raw fMRI data (i.e. representative time series of each anatomic region) and the clustering of the arc weights of mesh networks (MADs). Note that, clustering the MADs increases the rand index from $68\%$ to $84\%$. This substantial improvement shows that connectivity patterns are much more informative then the average voxel time-series.

Our next analysis involves the representation power of individual frequency sub-bands, where we examine the performance of each sub-band in detecting MADs among anatomic regions for the given tasks. This may also be translated as the amount of complementary information each sub-band carries in regard to the functional connectivity of the given cognitive states. For further comparison, we cluster the data after attaining the MADs at each sub-band (Section \ref{sec:hmmns}) and also after the deep learning architecture (Section \ref{sec:dl}). The results are stated in Table \ref{tab:table2}. The high rand indices for all individual sub-bands confirm the benefits of analyzing the fMRI signals in multiple time-resolutions as it shows that each sub-band carries important information in regard to the mesh network arc weights and the connectivity patterns learned at the output of stacked de-noising auto-encoders. This leads to a clustering performance between the range of $68-86\%$ across all sub-bands. Note that, the sub-bands \bf{A5} \normalfont to \bf{A11}\normalfont, \bf{D5} \normalfont to \bf{D7}\normalfont~and \bf{D9} \normalfont to \bf{D11}\normalfont~show relatively higher performances indicating that these frequency bands are more informative then the rest. 

In order to boost the RI and ARI values given in Table \ref{tab:table2}, we fuse the learned connectivity patterns based on a combination of sub-bands to obtain a better representation. In our last set of clustering analyses, we examine the clustering performance by ensembling multiple sub-bands. RI and ARI values for the ensembled sub-bands, given in Table \ref{tab:table3} point to a substantial increase compared to the best single sub-band clustering performance of $86\%$ at sub-band A10 to a performance of $93\%$ for the fusion of all sub-bands. This shows that not only the brain networks constructed at multiple time-resolutions provide complementary information for the clustering algorithm but that the proposed deep architecture is capable of detecting distinct connectivity patterns in the brain for a given cognitive task, independent of subjects.

The rather high ARI values in Table \ref{tab:table3} confirm that utilizing the complementary information gained from different time-resolutions result in clusters with relatively low within-cluster variances and high between-cluster variances. This claim is backed by the high ARI values that result from combining the information from different sub-bands before clustering. Further, by increasing the number of subjects to $200$ in our data set, and by fusing the brain networks obtained from the entire $23$ sub-bands and clustering their connectivity pattern extracted by the SDAE platform, we are able to achieve the performance of $93\%$ RI and $71\%$ ARI. This experiment shows that increasing the number of subjects does not decrease the clustering performance.
\begin{figure*}[h]
	\centering
	\begin{subfigure}
		\centering
		\includegraphics[scale=0.2539]{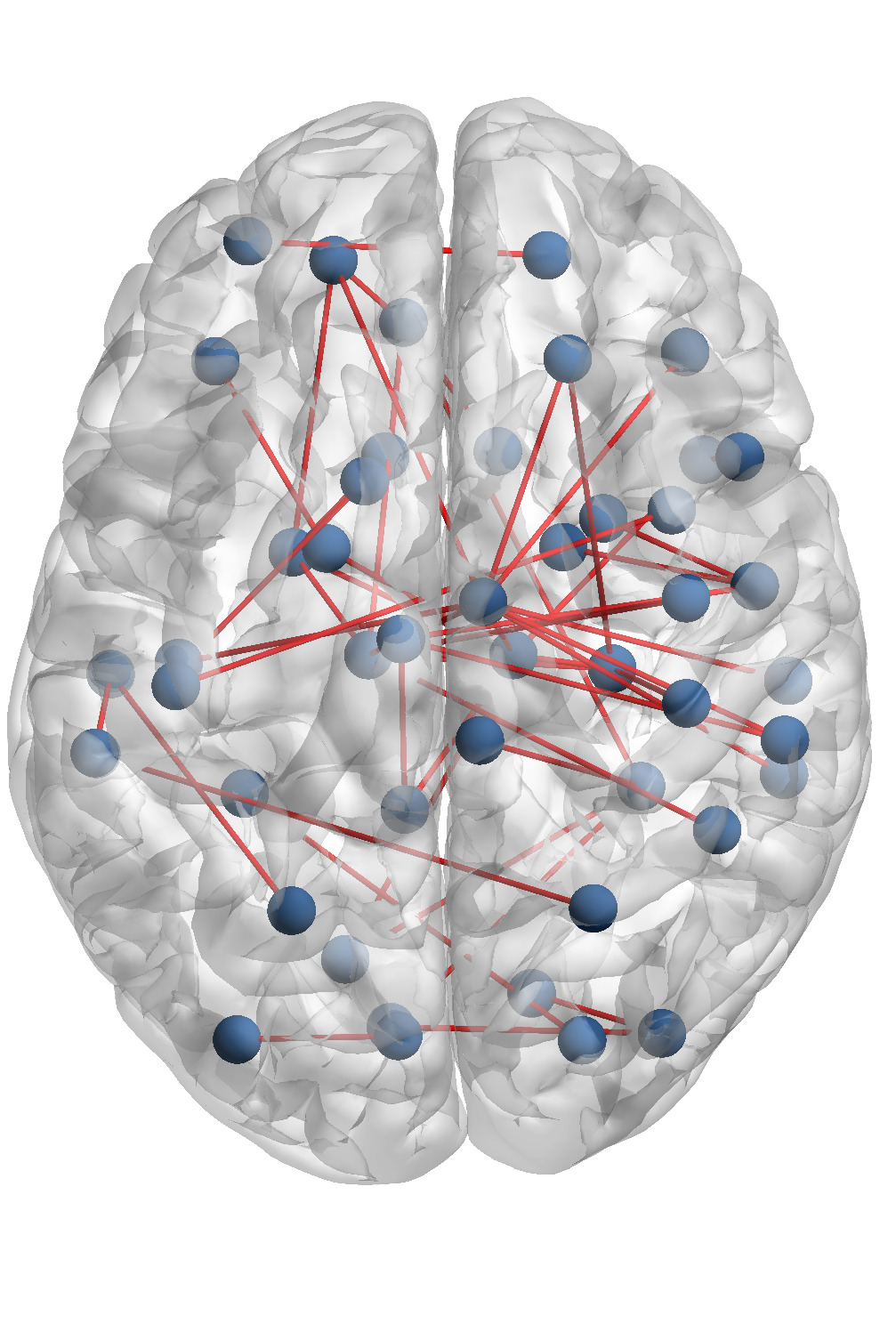} 
		\includegraphics[scale=0.2539]{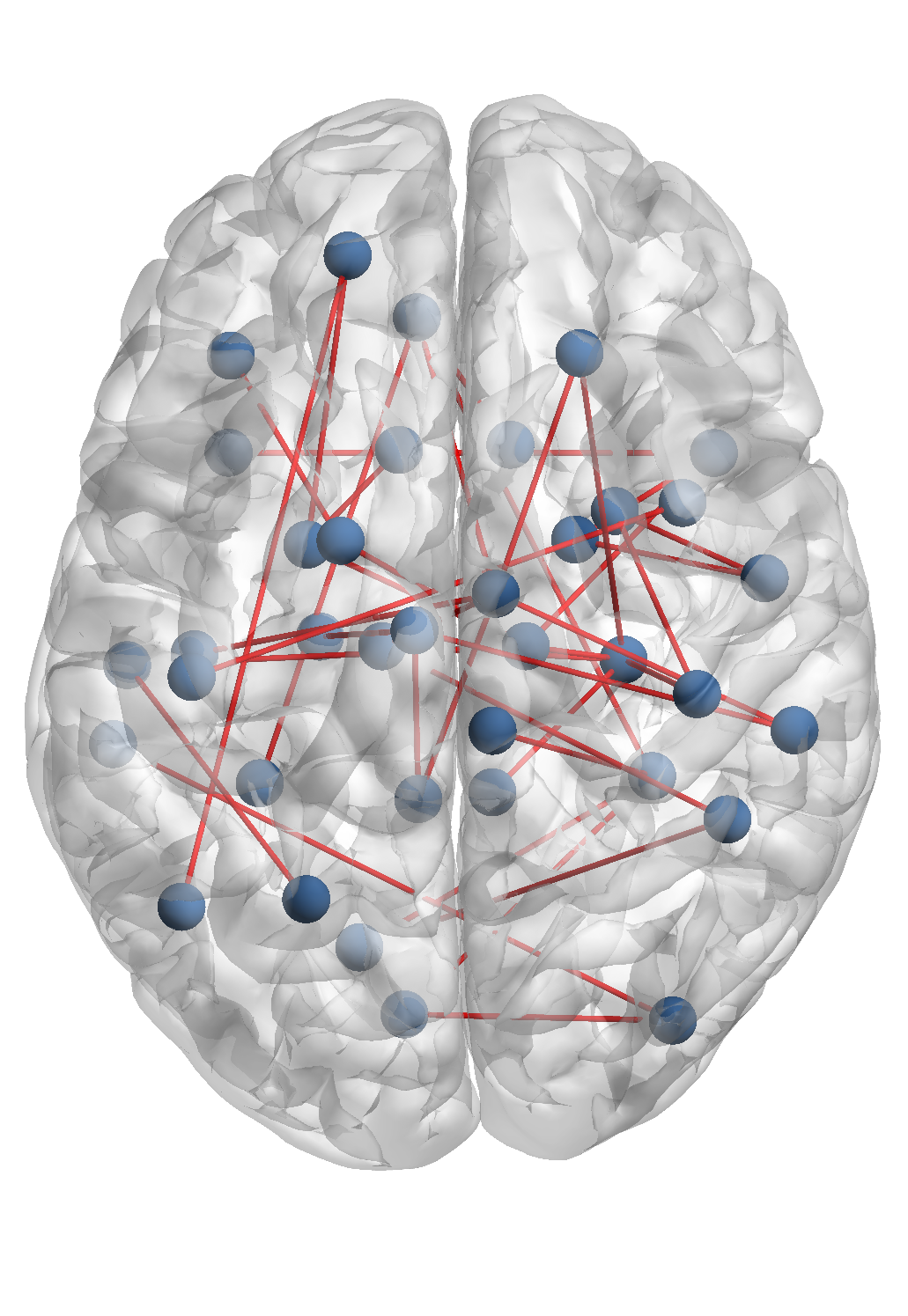} 
		\includegraphics[scale=0.2539]{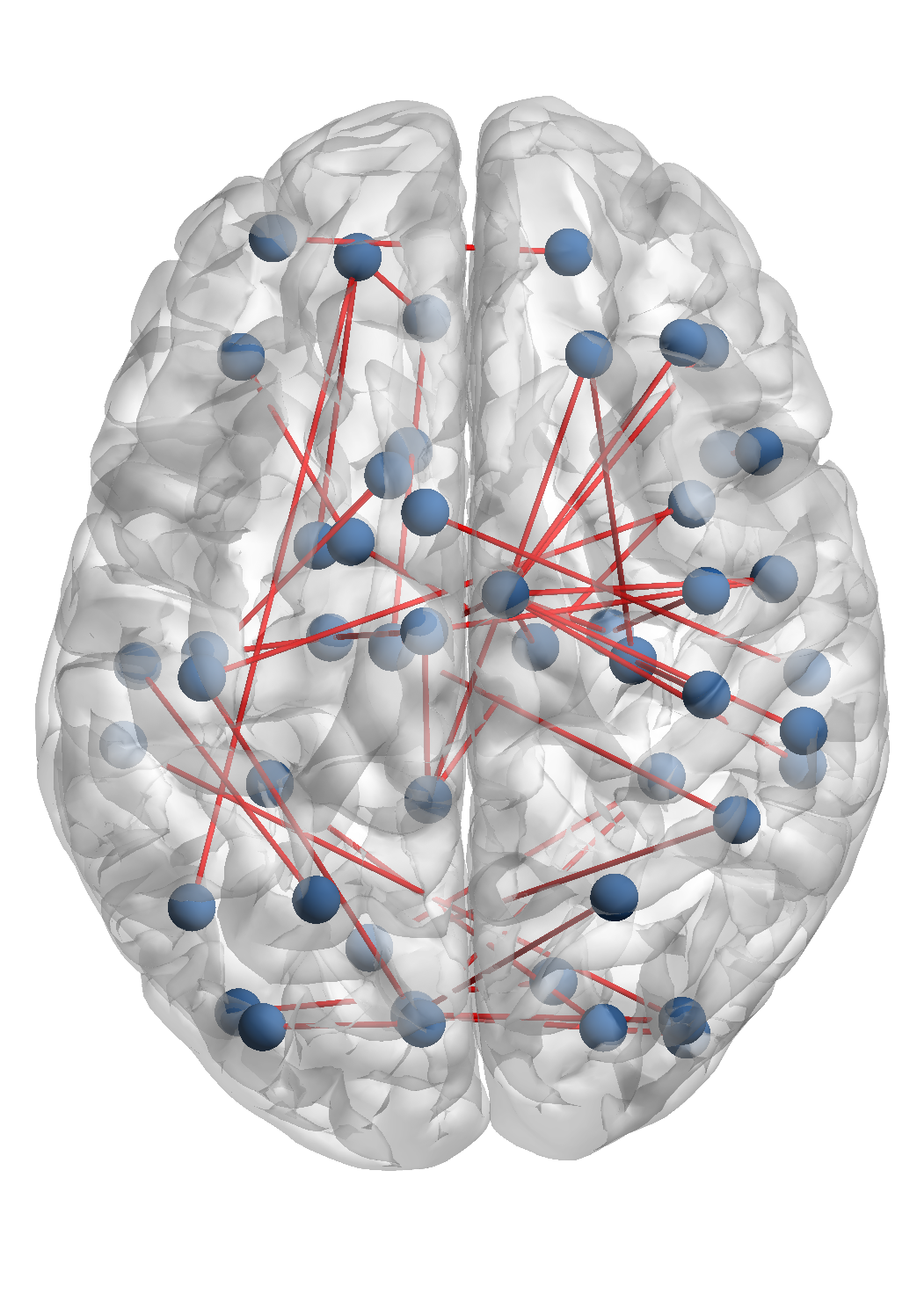} 
		\includegraphics[scale=0.2539]{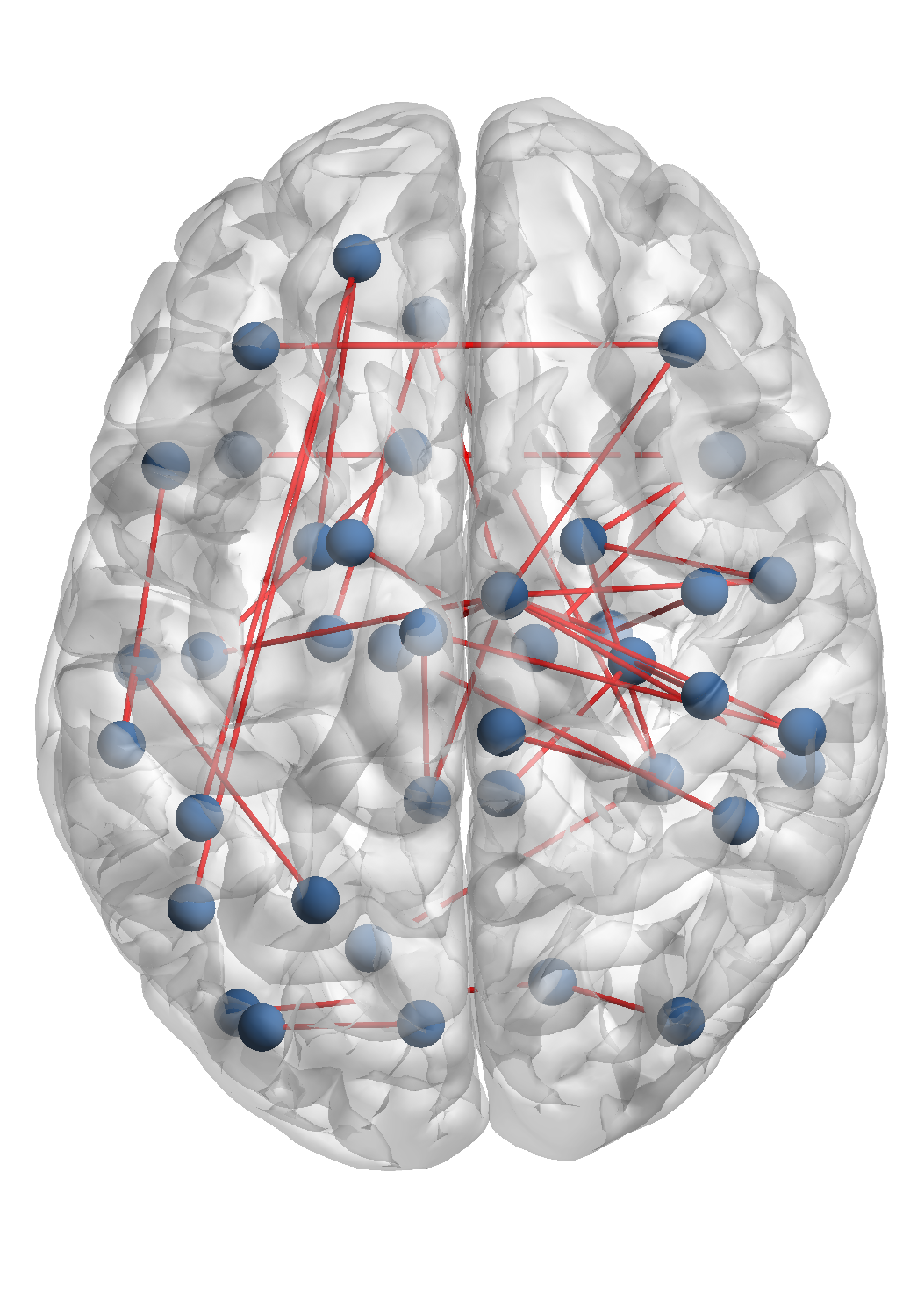} 
		\includegraphics[scale=0.2539]{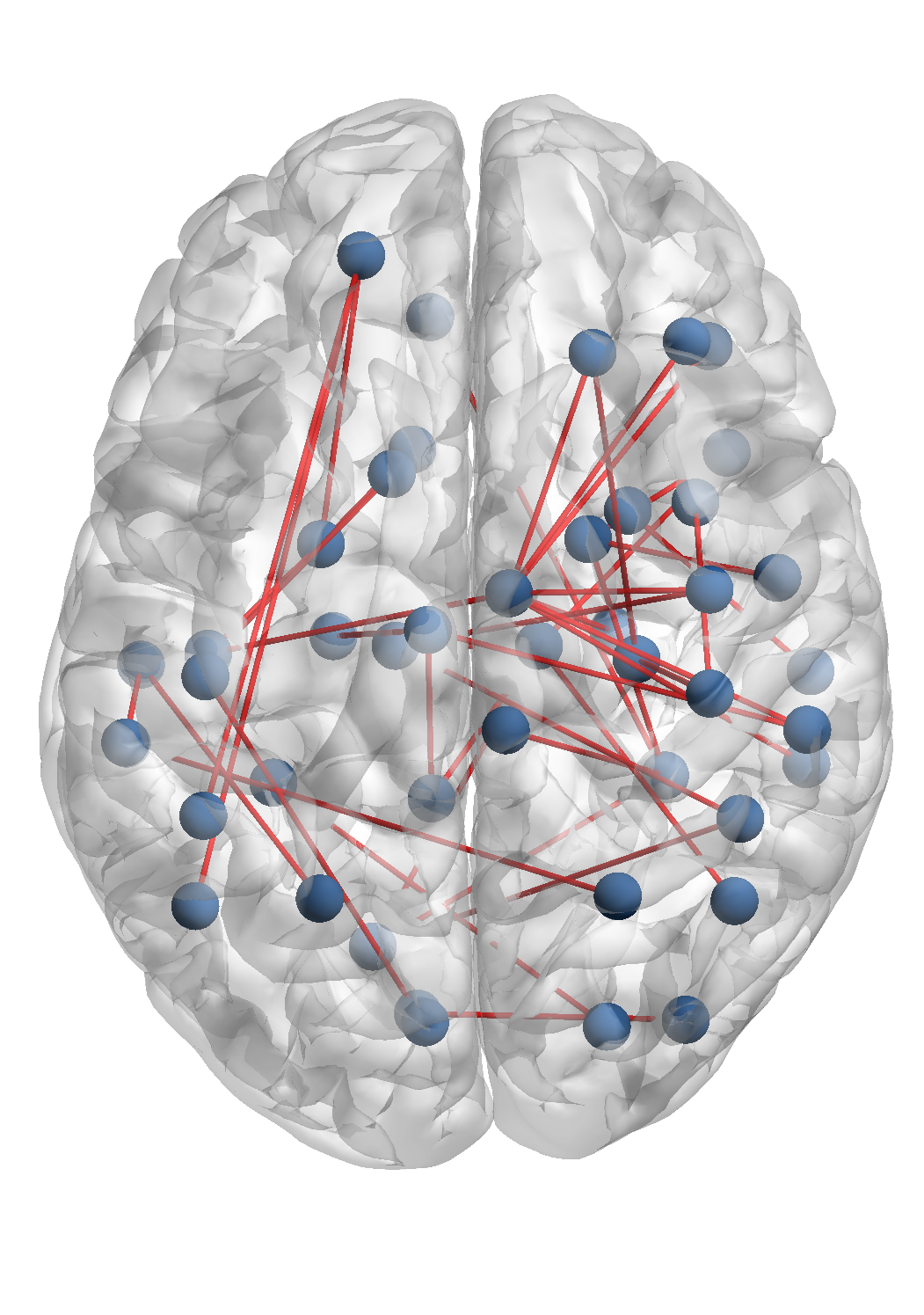} 
		\includegraphics[scale=0.2539]{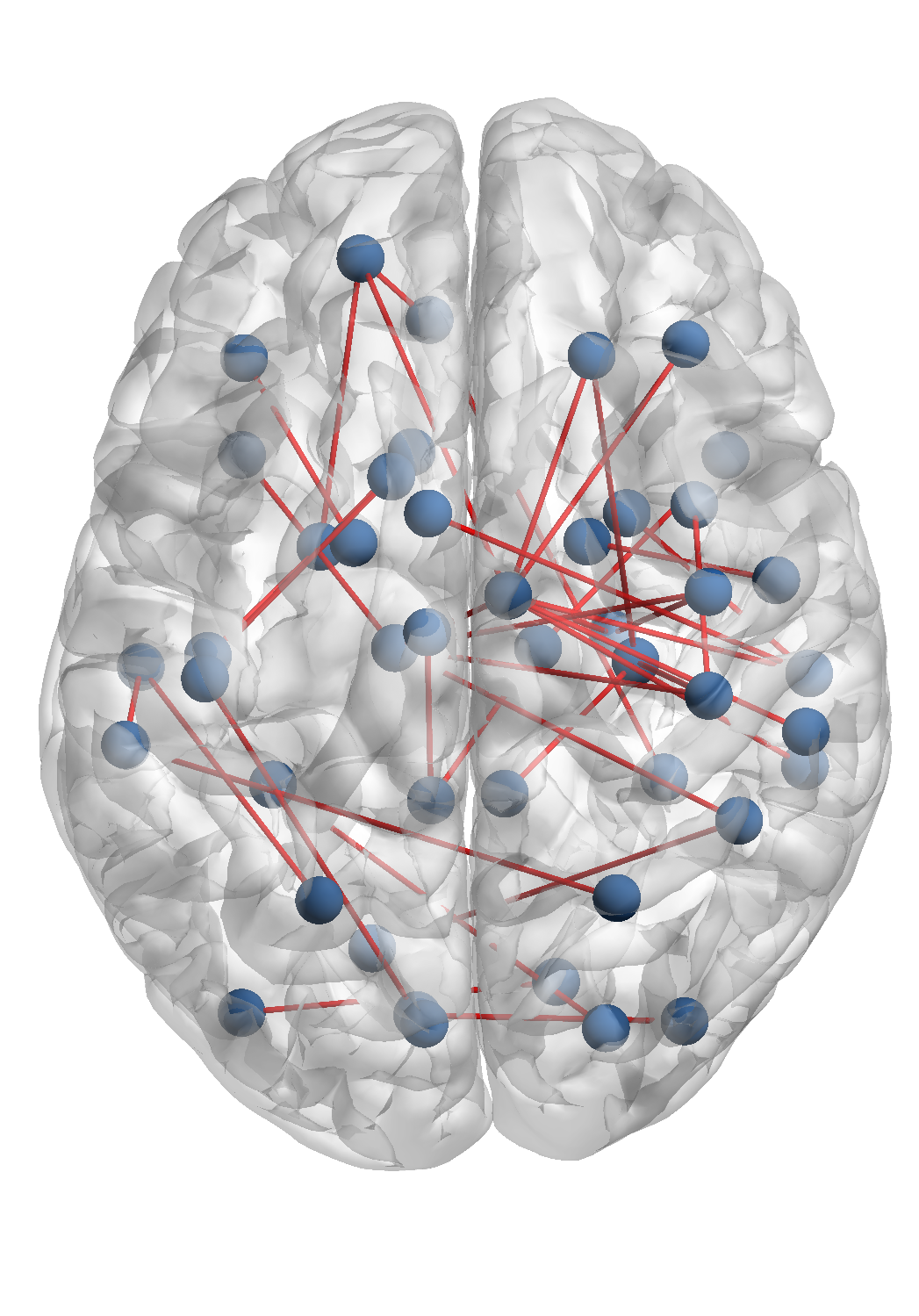} 
		\includegraphics[scale=0.2539]{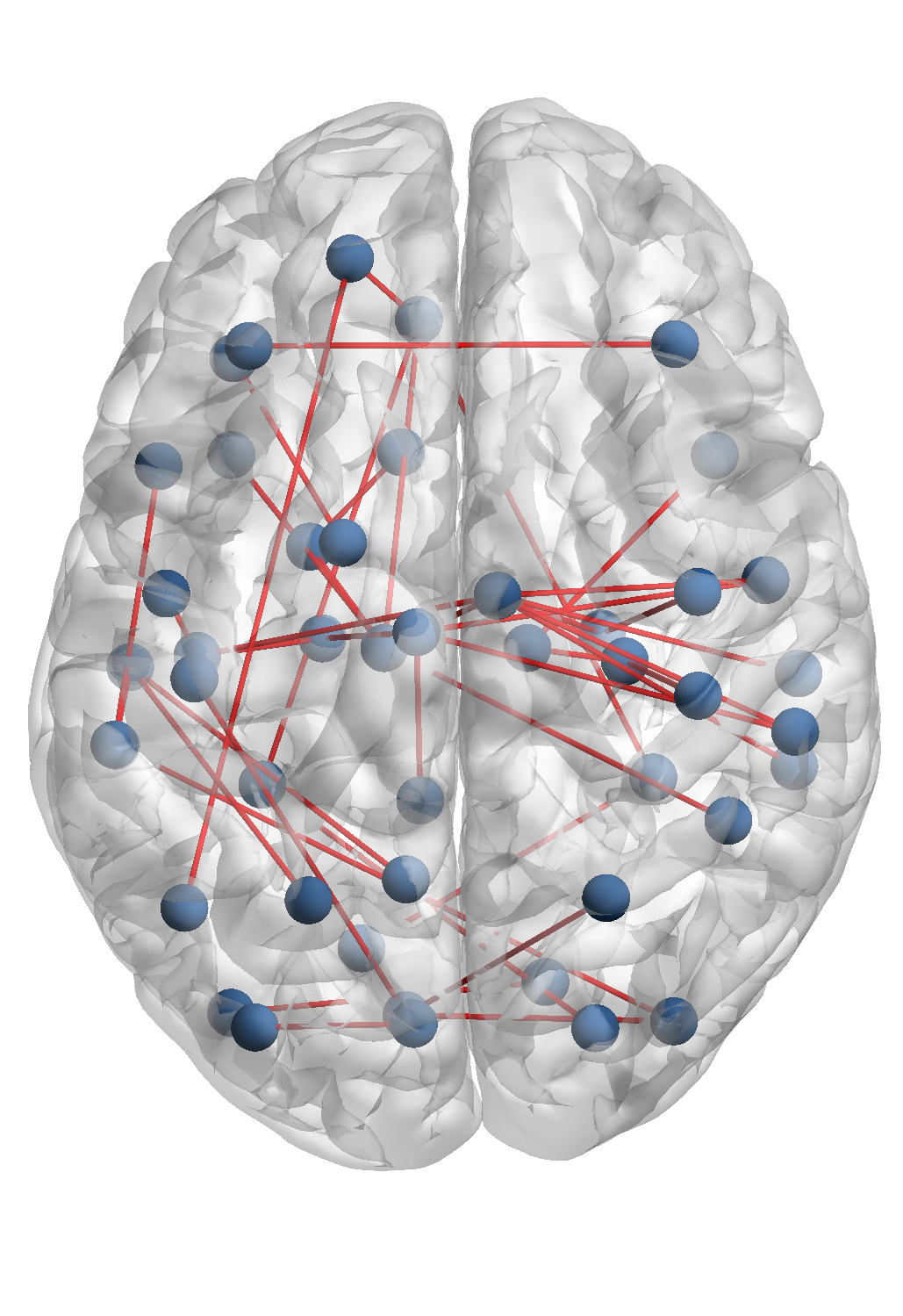} 
	\end{subfigure}
	\begin{subfigure}
		\centering
		\includegraphics[scale=0.2539]{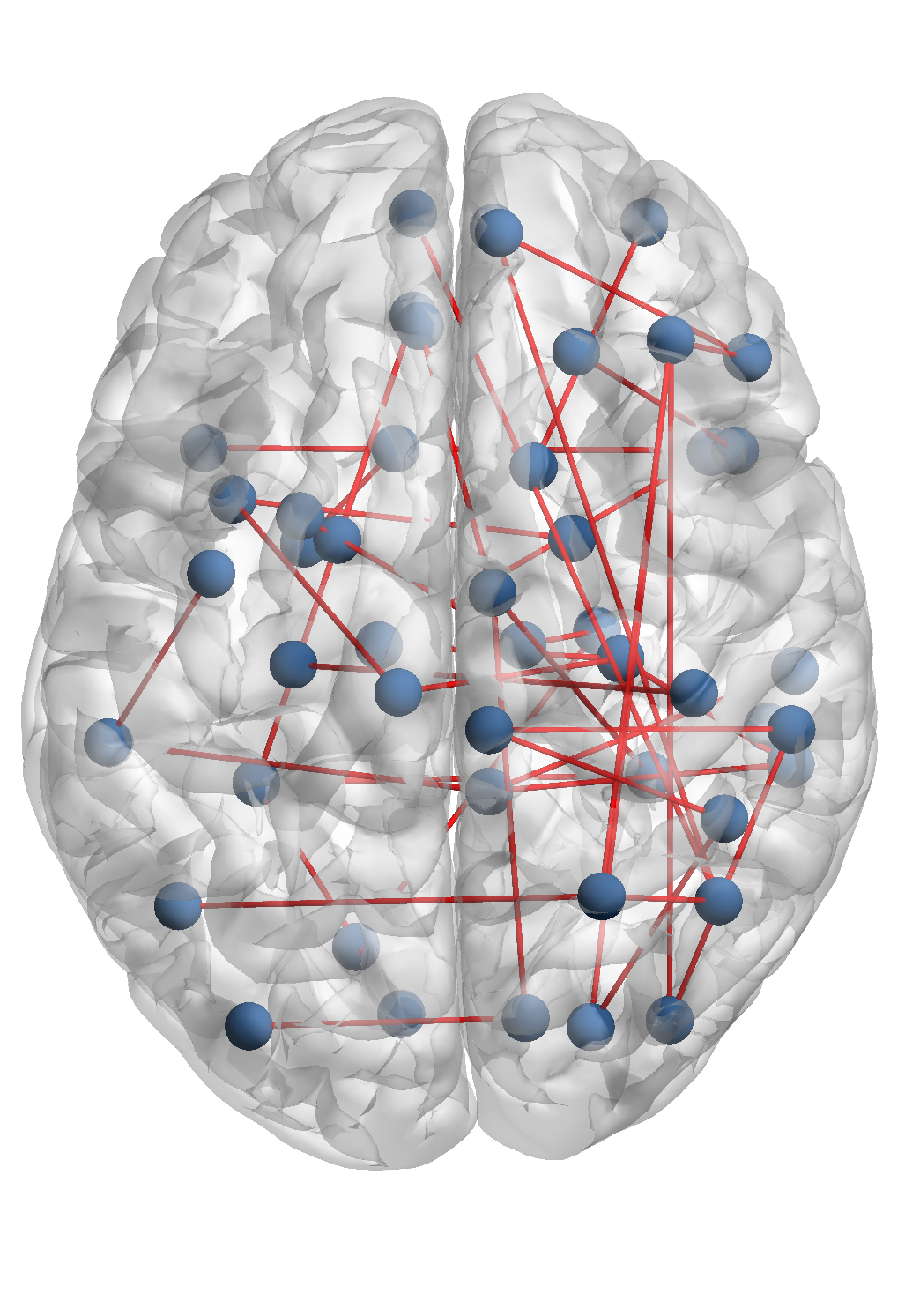}
		\includegraphics[scale=0.2539]{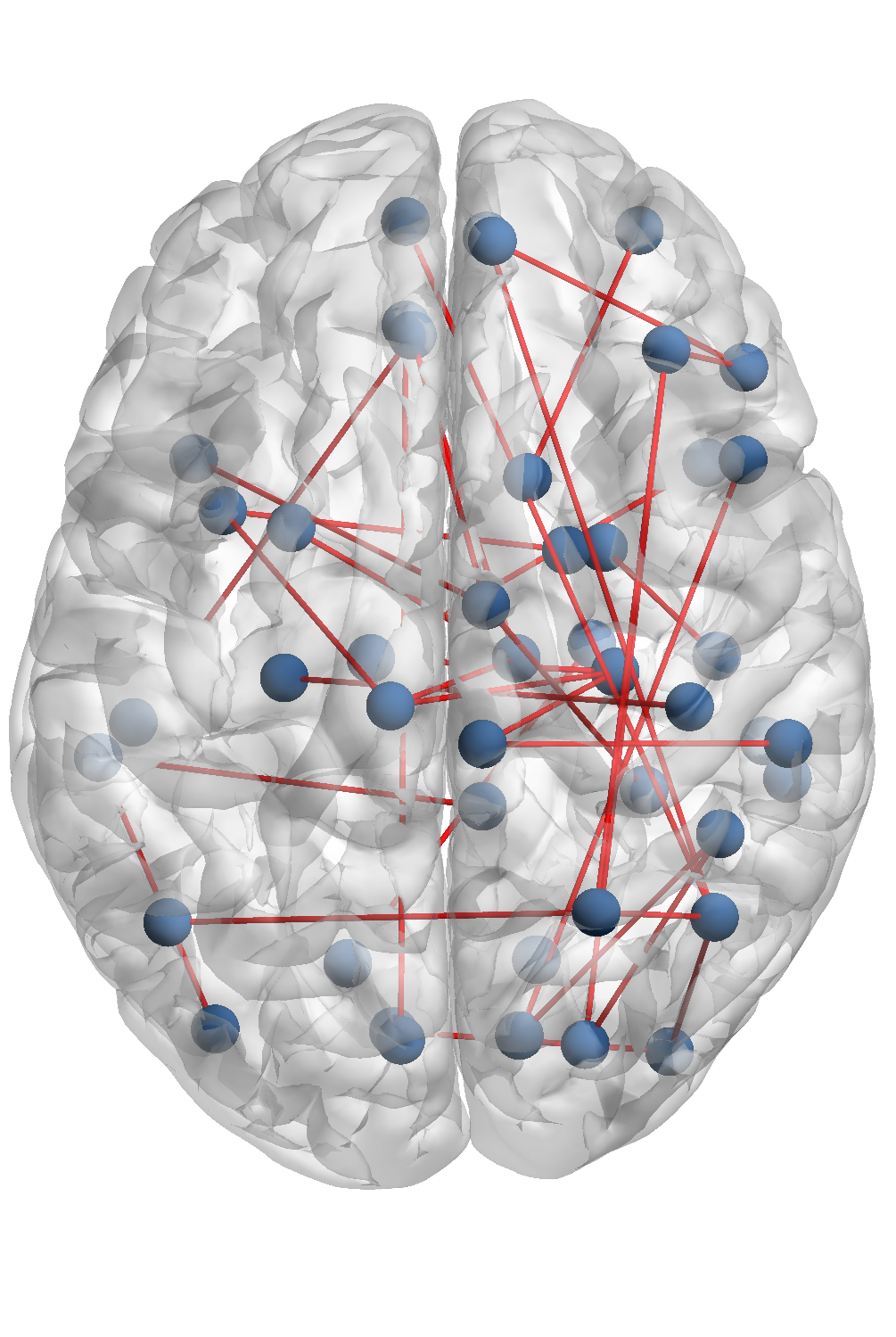}
		\includegraphics[scale=0.2539]{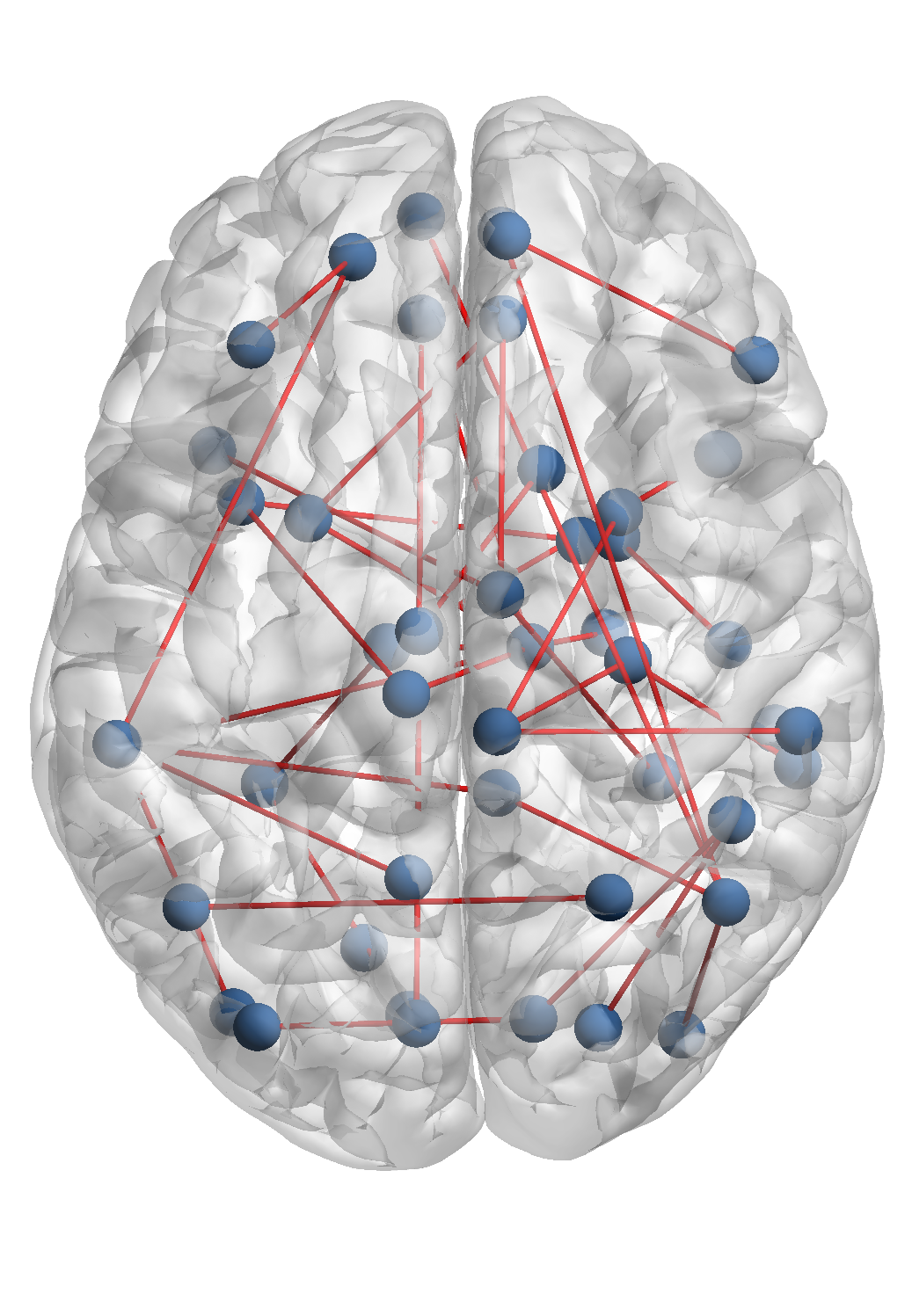}
		\includegraphics[scale=0.2539]{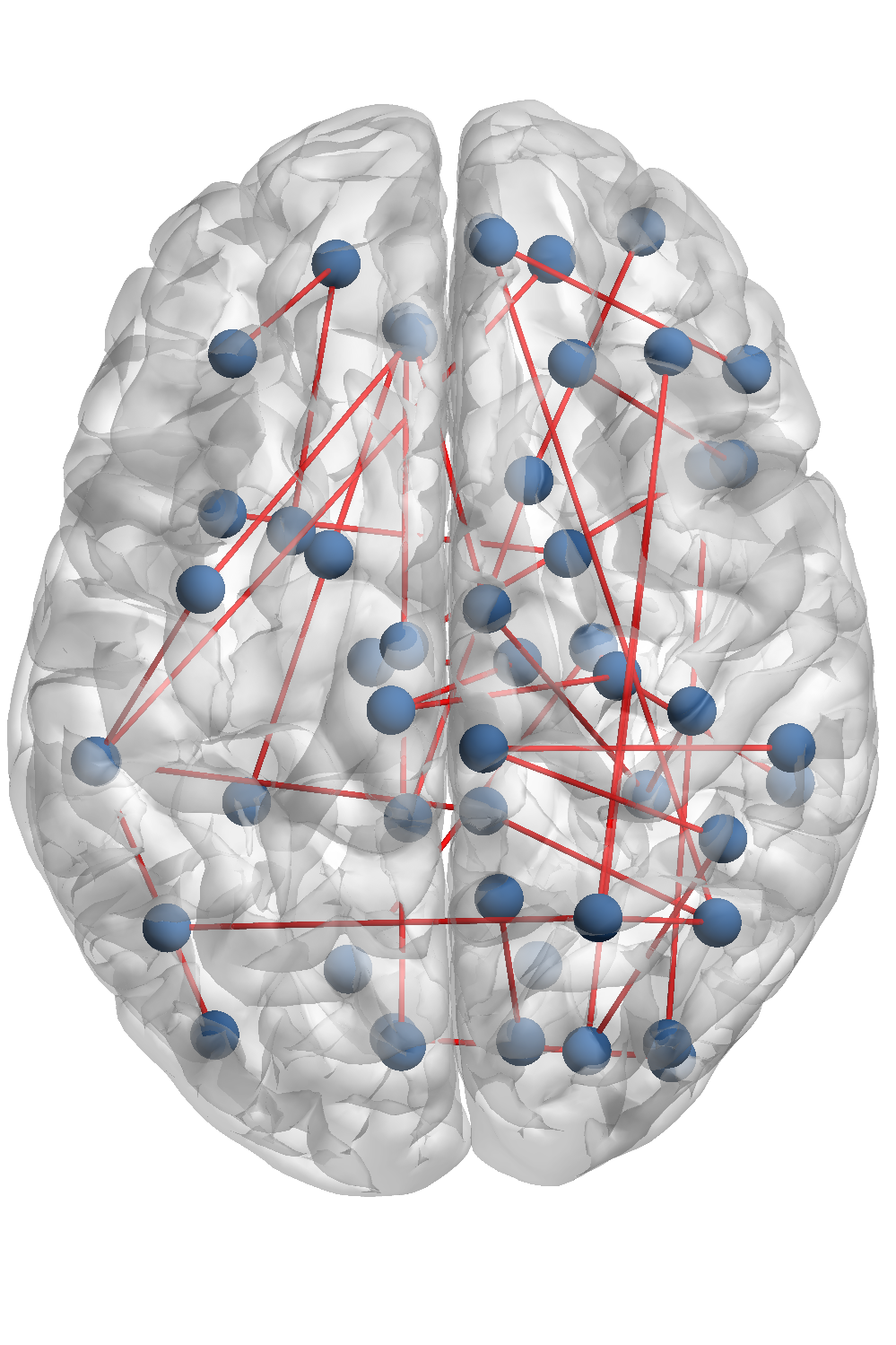}
		\includegraphics[scale=0.2539]{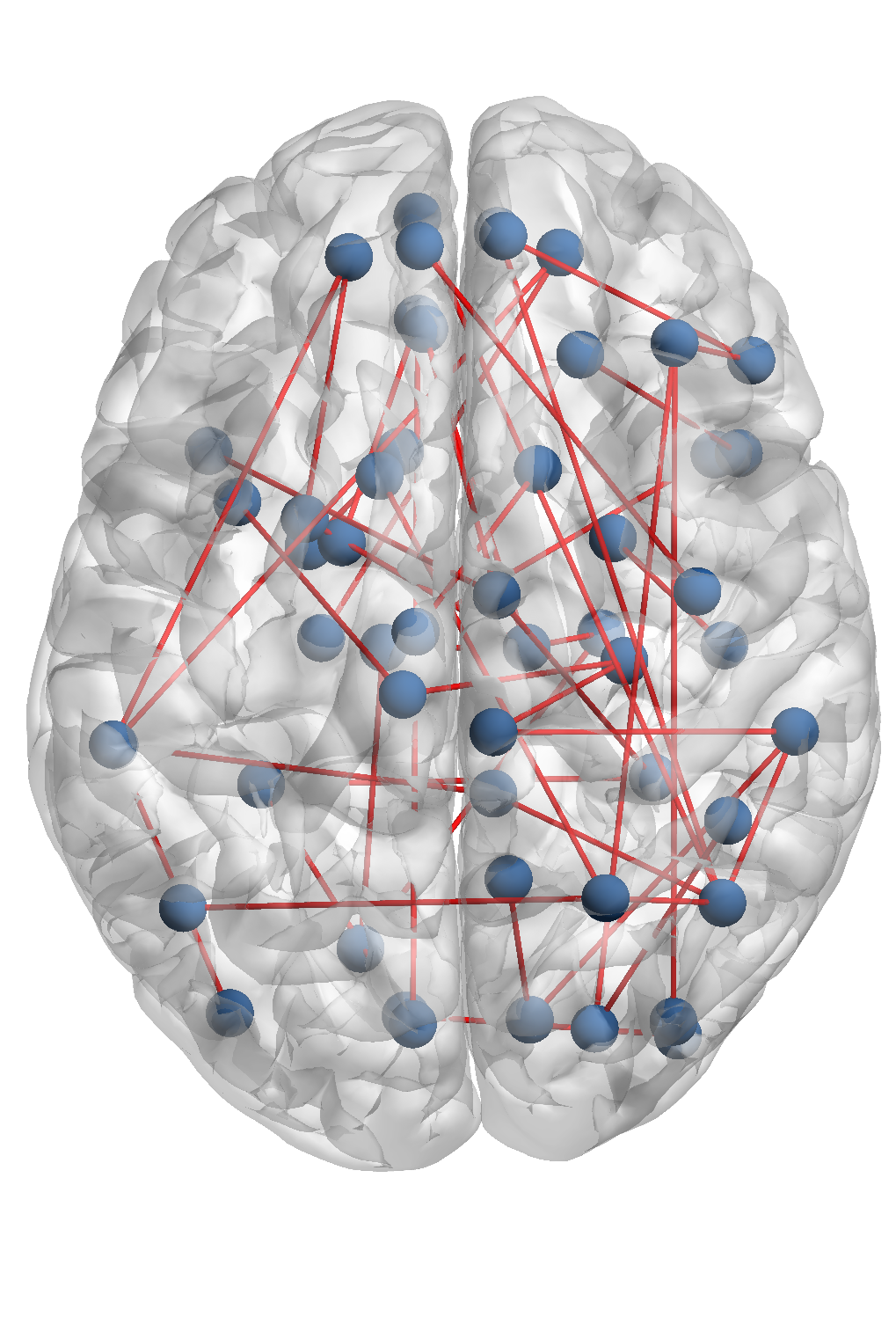}
		\includegraphics[scale=0.2539]{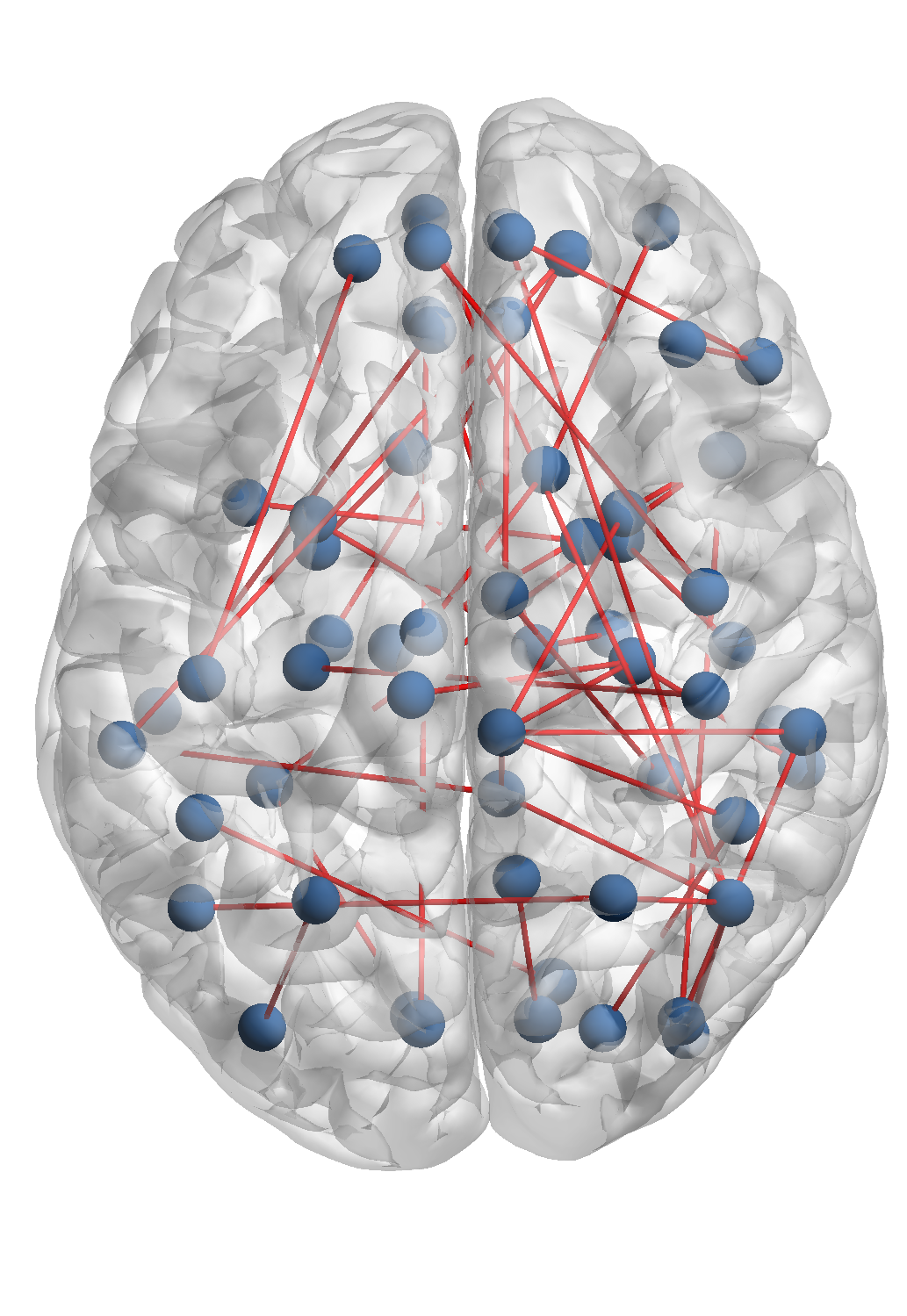}
		\includegraphics[scale=0.2539]{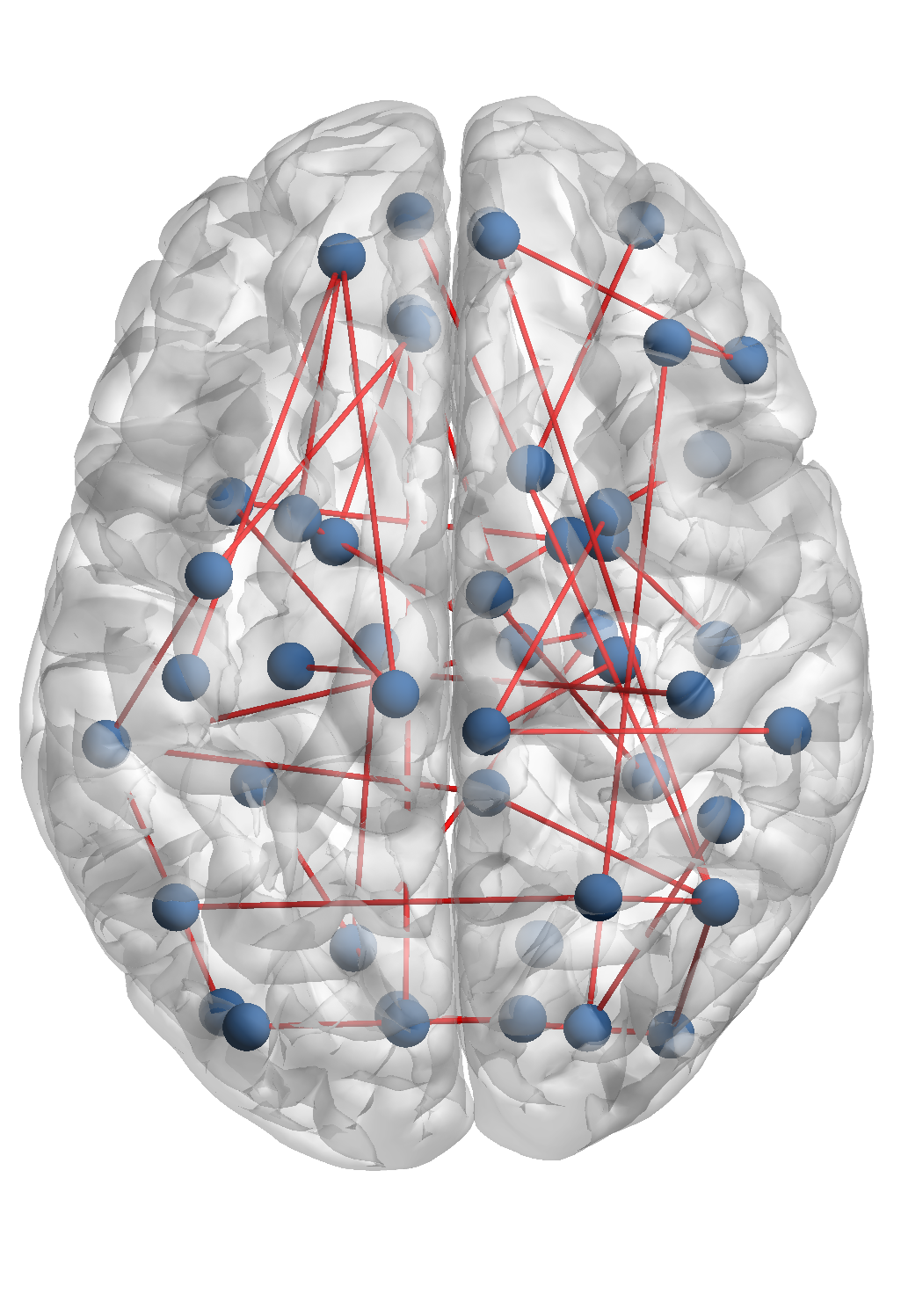}
	\end{subfigure}
	\begin{subfigure}
		\centering
		\includegraphics[scale=0.2539]{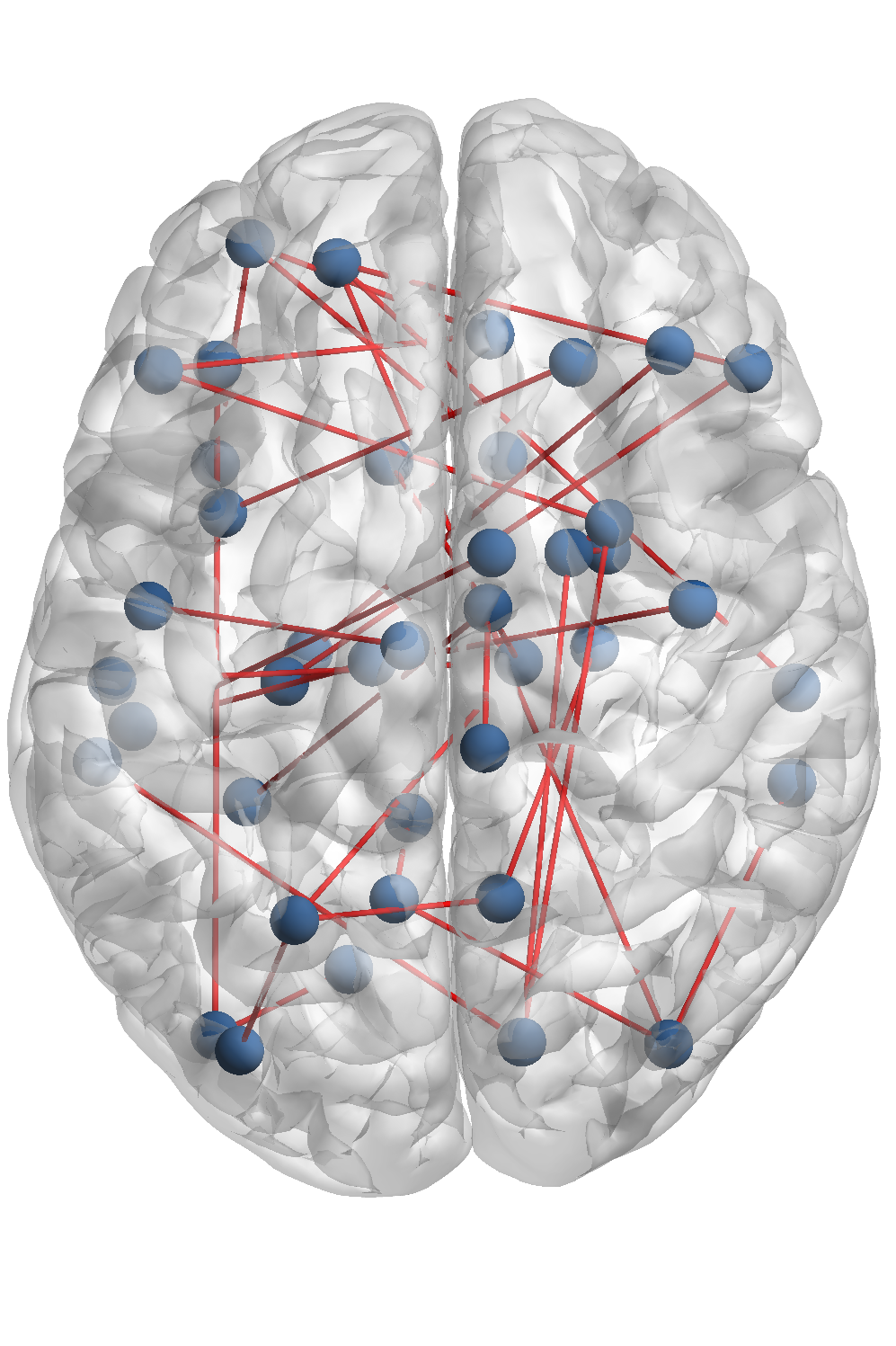}
		\includegraphics[scale=0.2539]{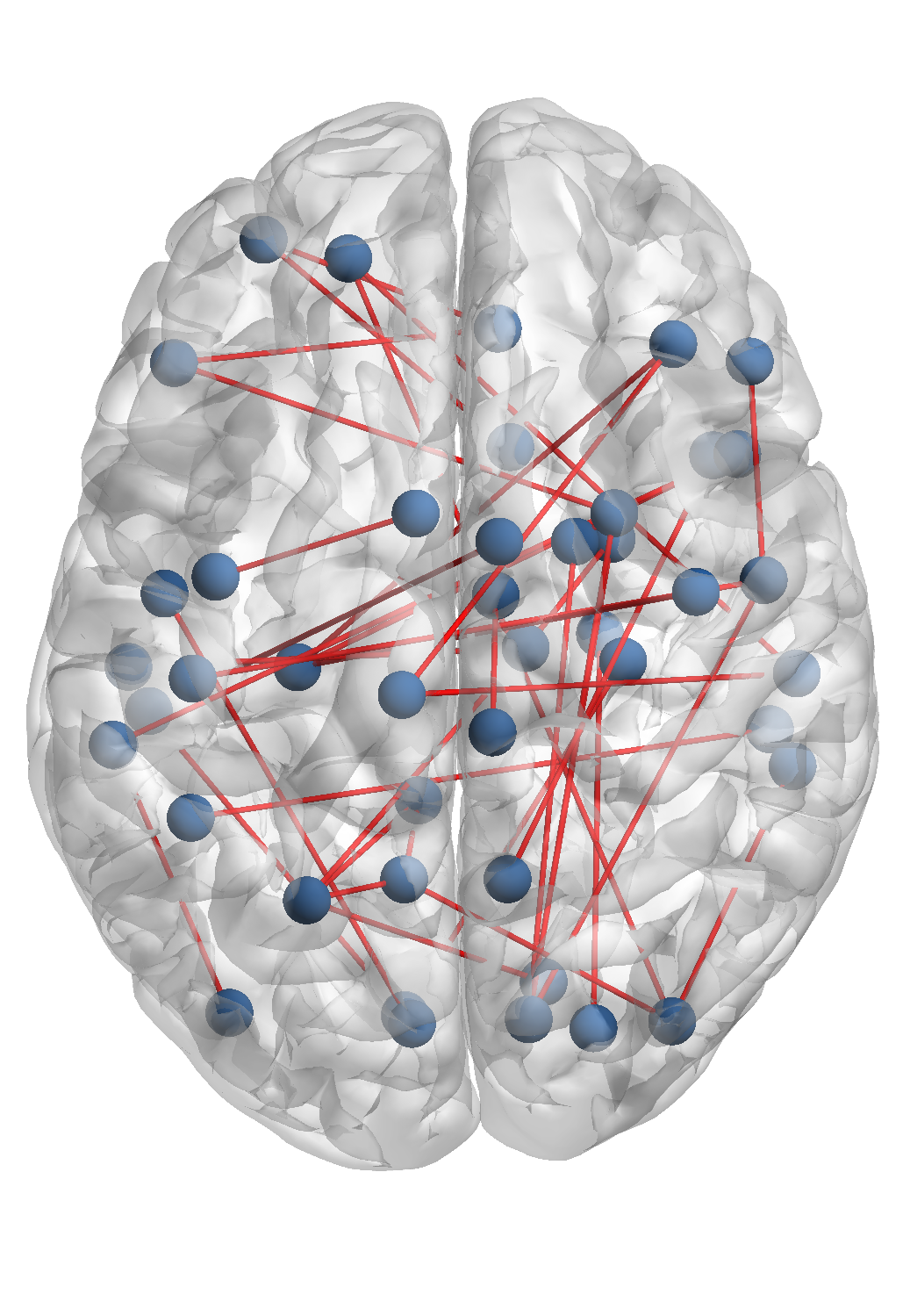}
		\includegraphics[scale=0.2539]{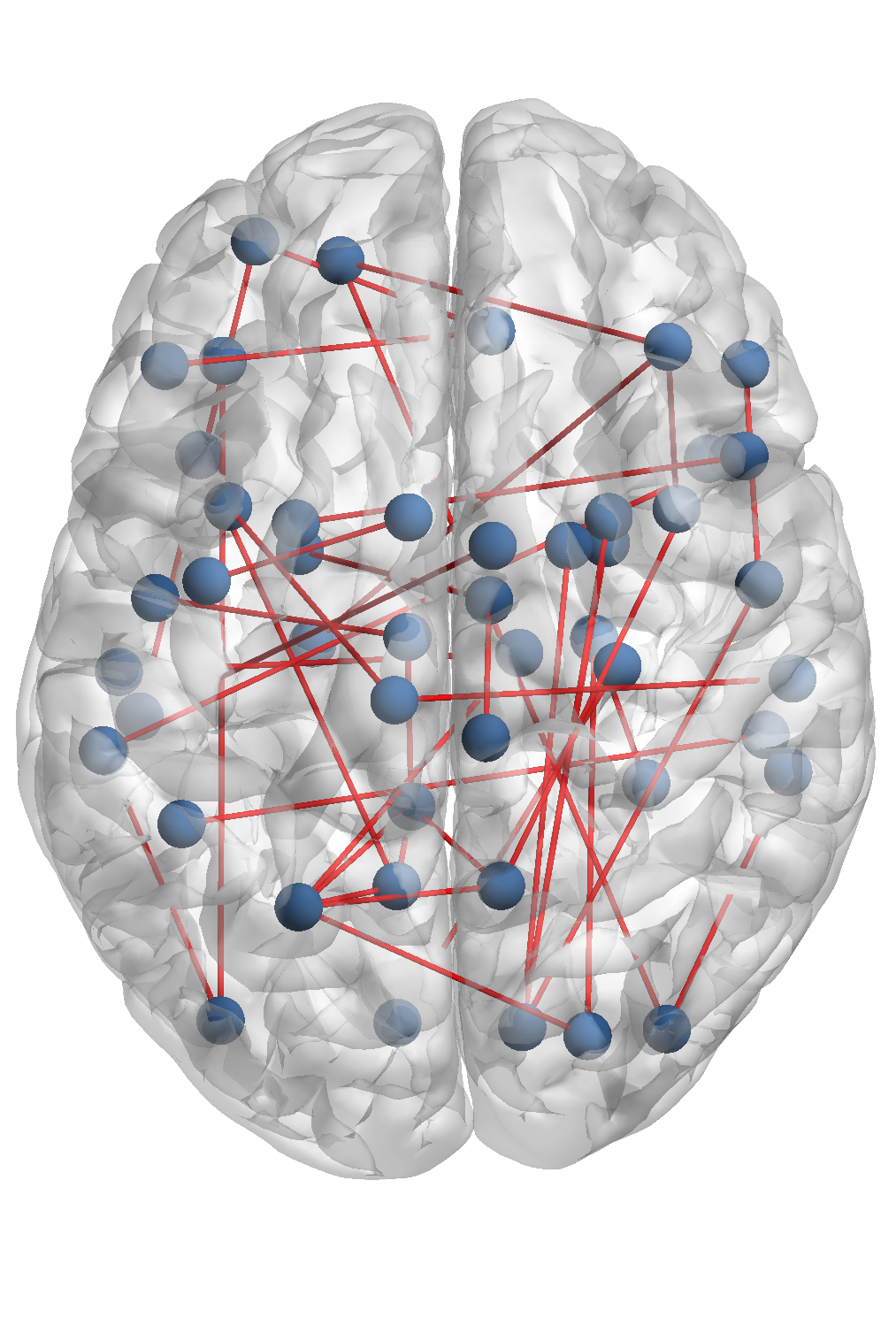}
		\includegraphics[scale=0.2539]{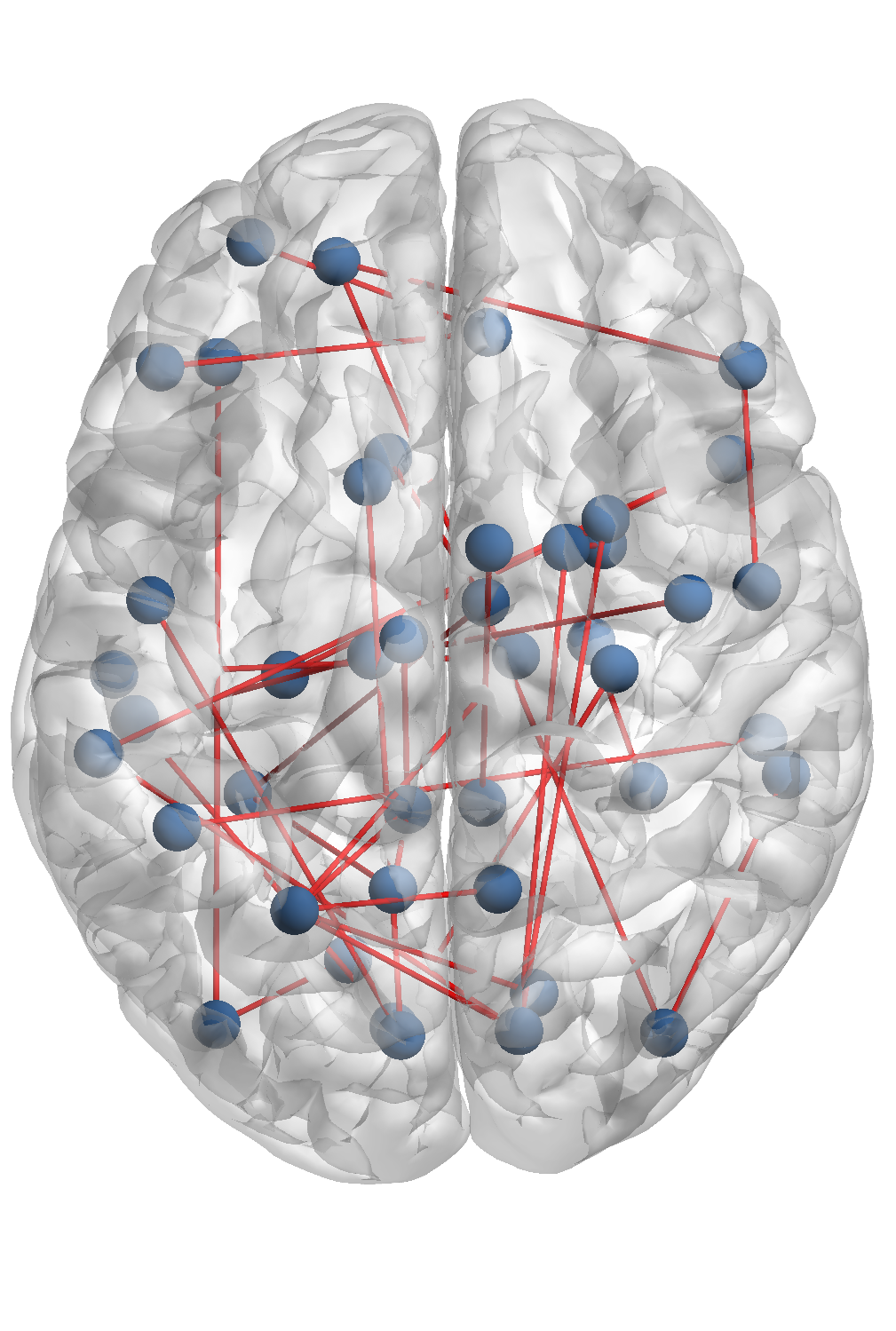}
		\includegraphics[scale=0.2539]{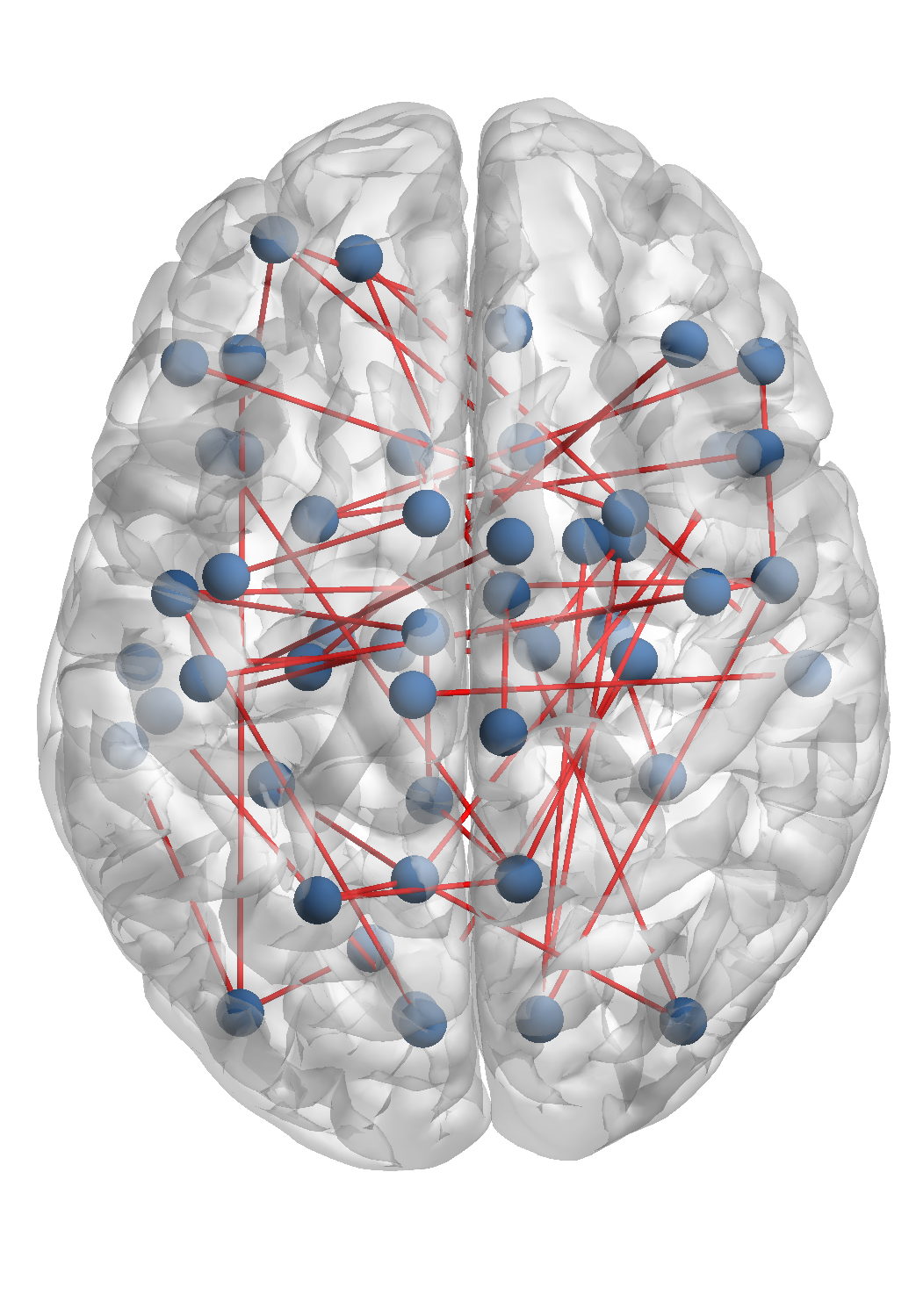}
		\includegraphics[scale=0.2539]{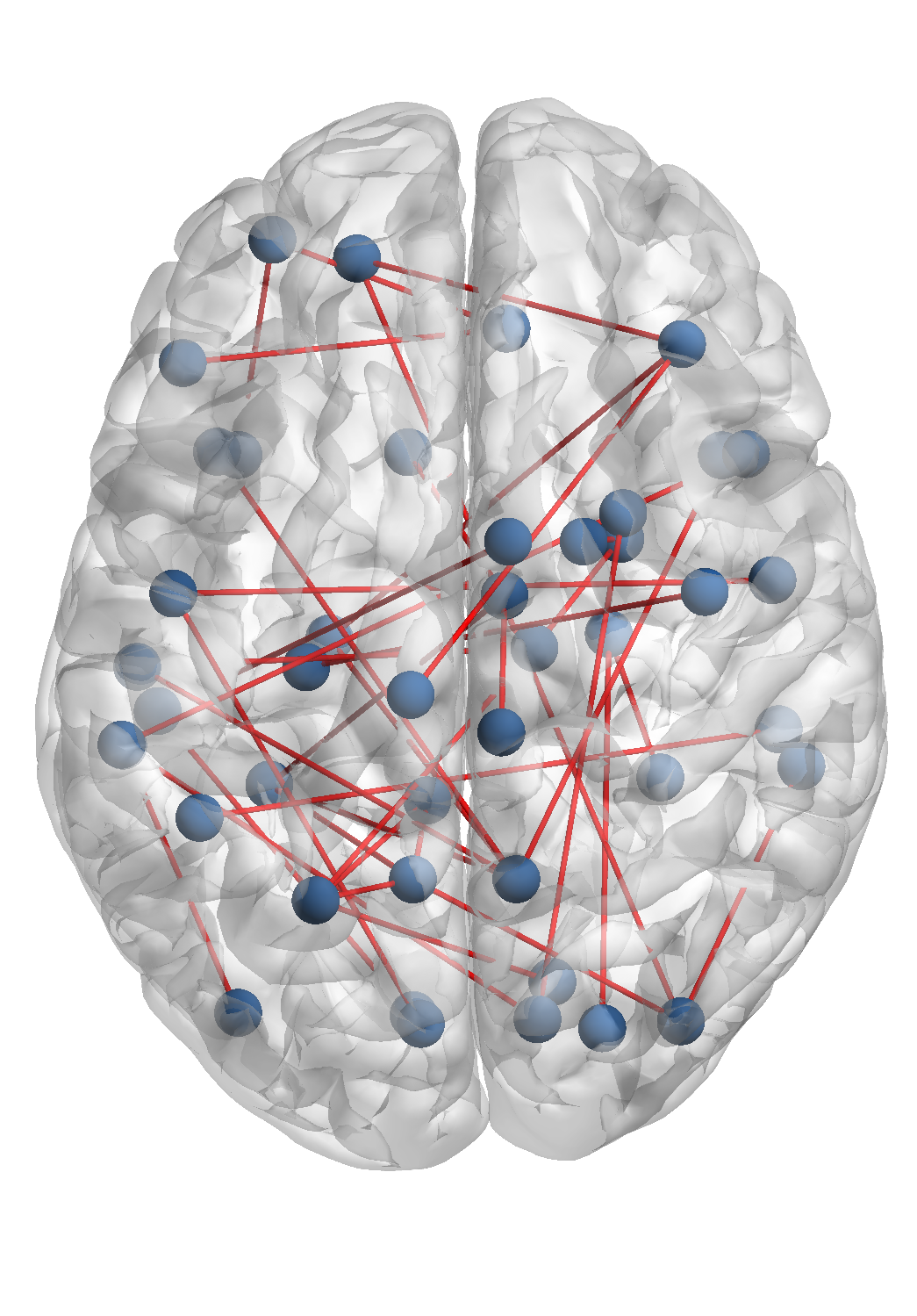}
		\includegraphics[scale=0.2539]{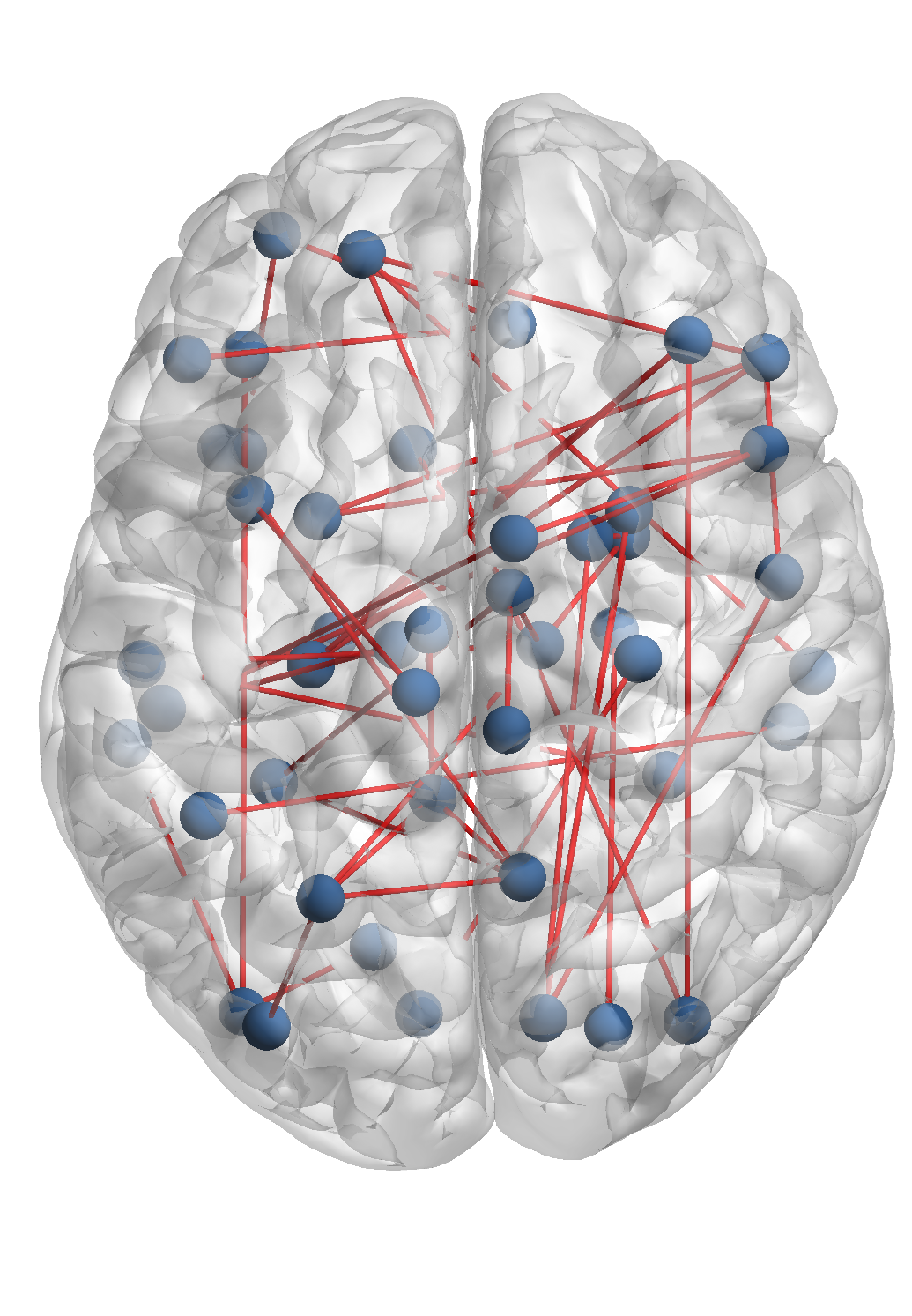}
	\end{subfigure}
	\caption{Mesh Networks of 3 Subjects: the top row shows subject 29, the middle row shows subject  40, bottom row shows subject 56. The mesh networks at each row from left to right indicate tasks in the following order: emotion, gambling, language, motor, relational, social and working-memory.}\label{fig:2}		
\end{figure*}
\begin{figure*}[h]
	\centering
	\includegraphics[scale=0.252]{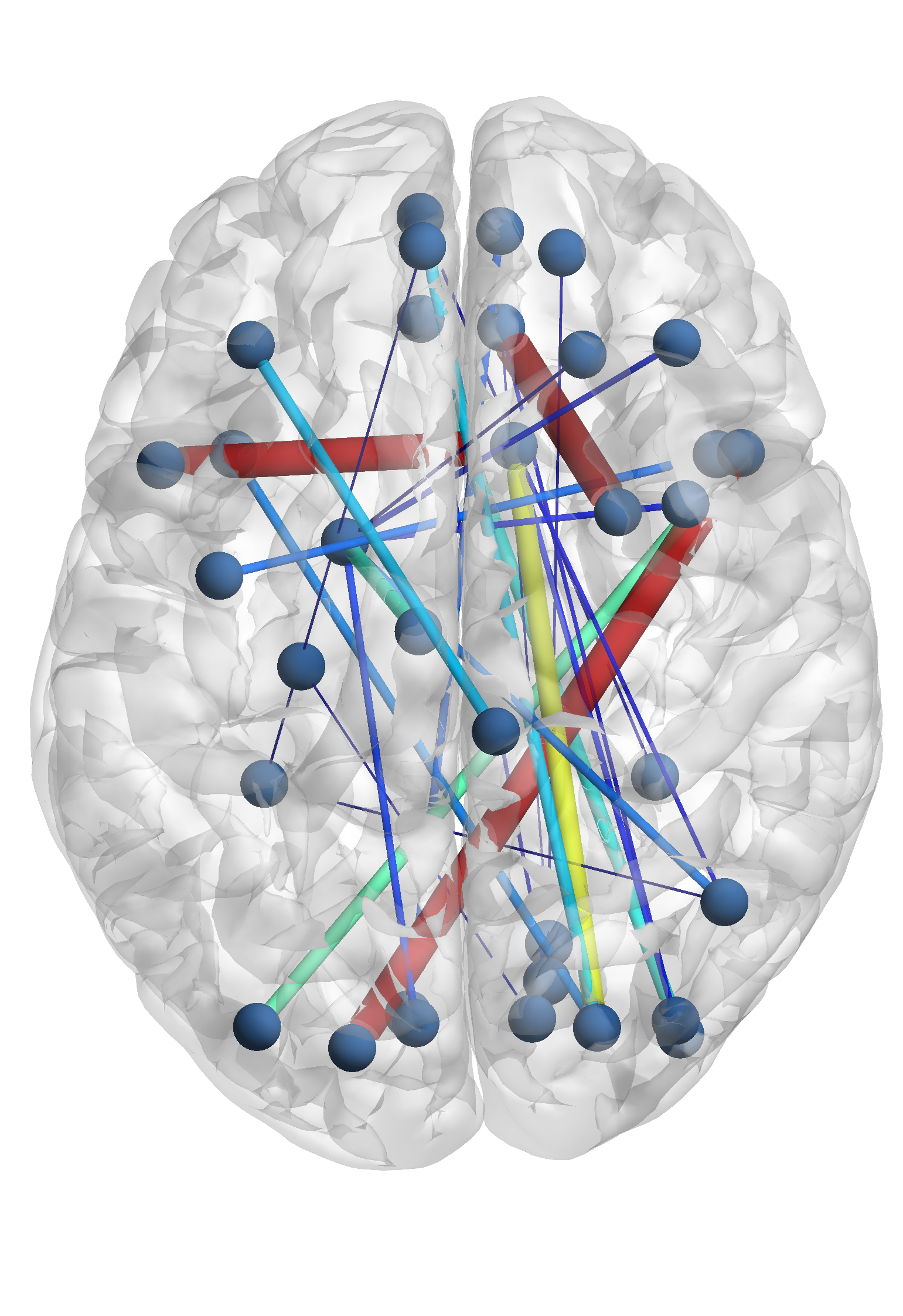}
	\includegraphics[scale=0.252]{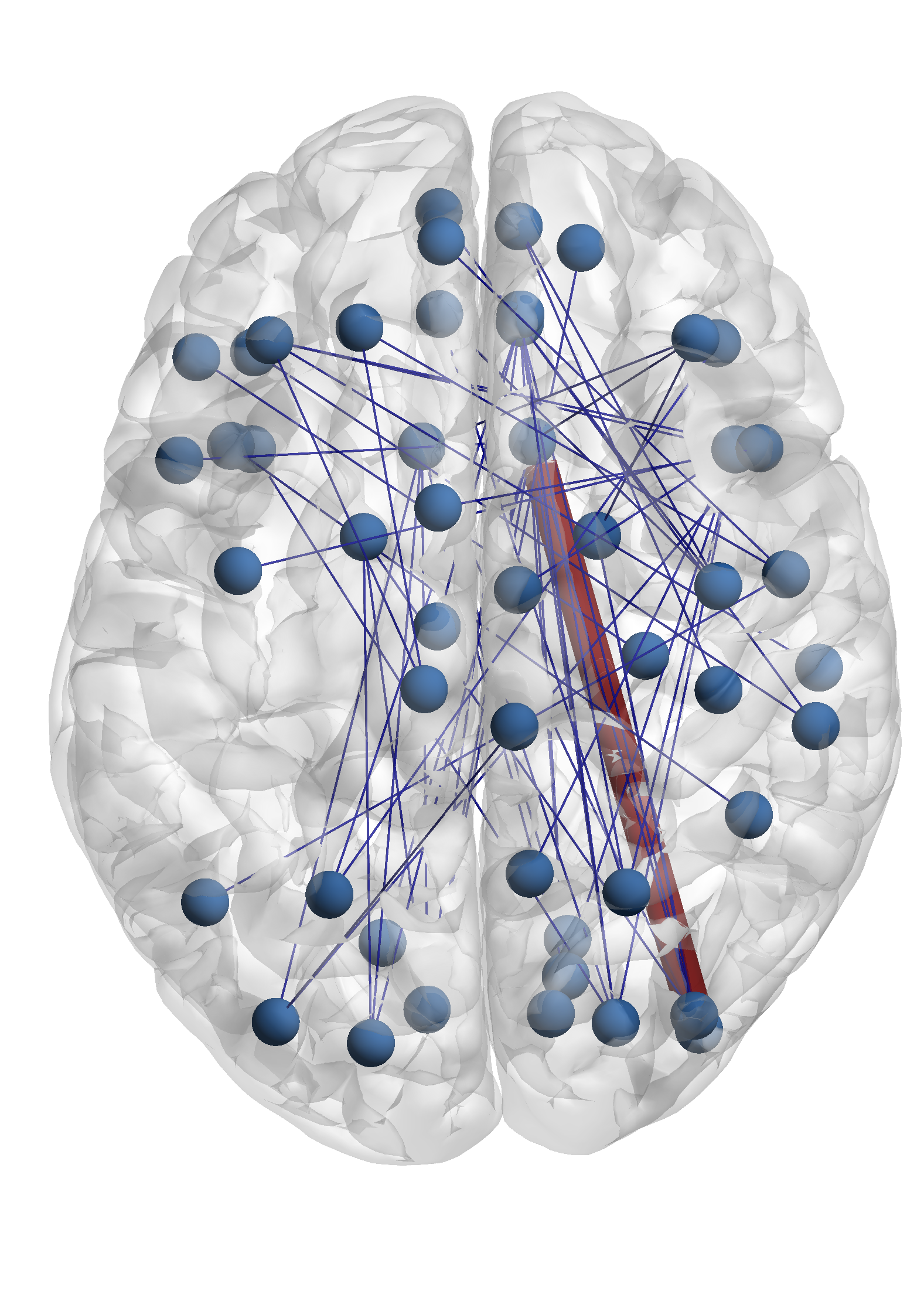}
	\includegraphics[scale=0.252]{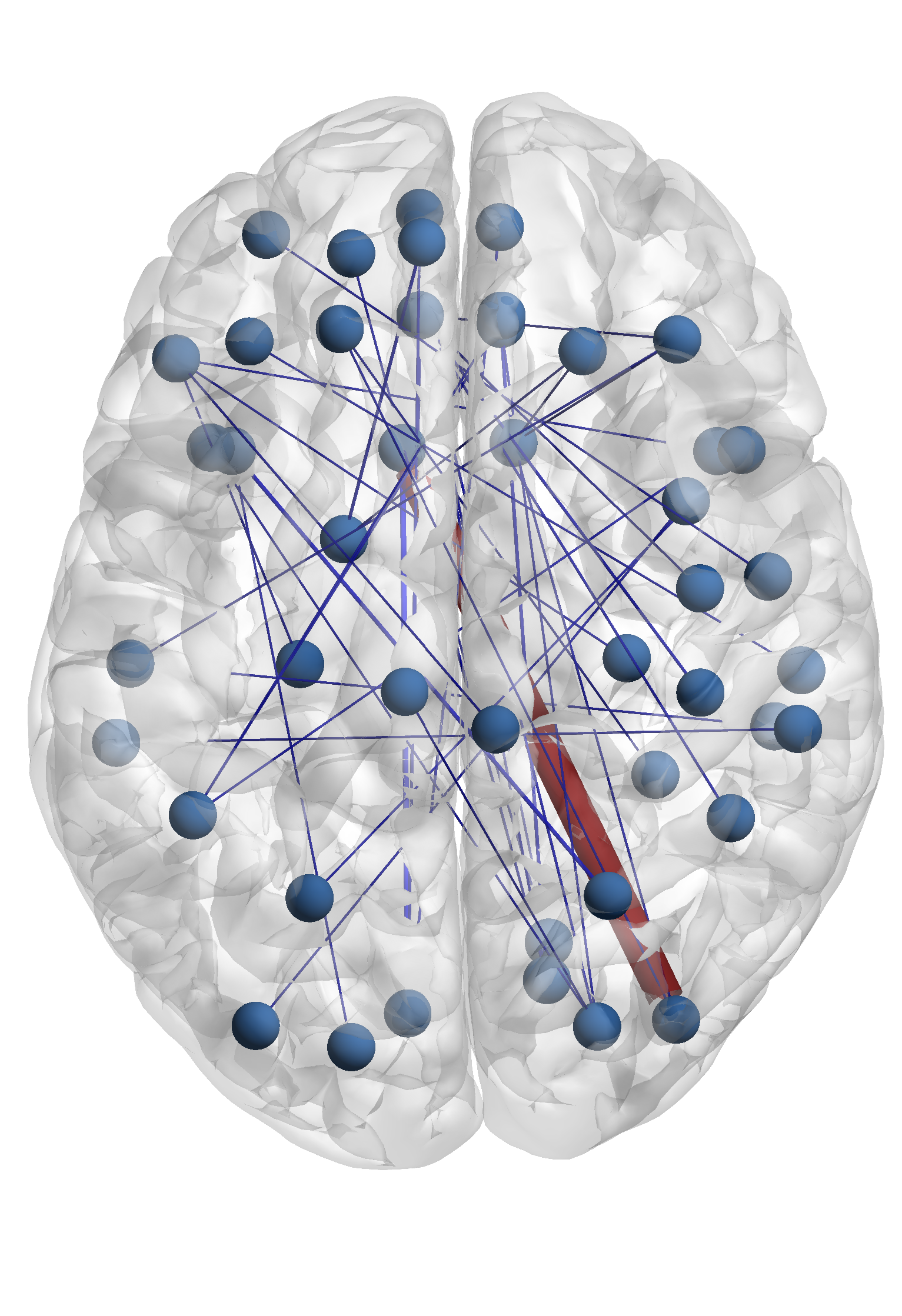}
	\includegraphics[scale=0.252]{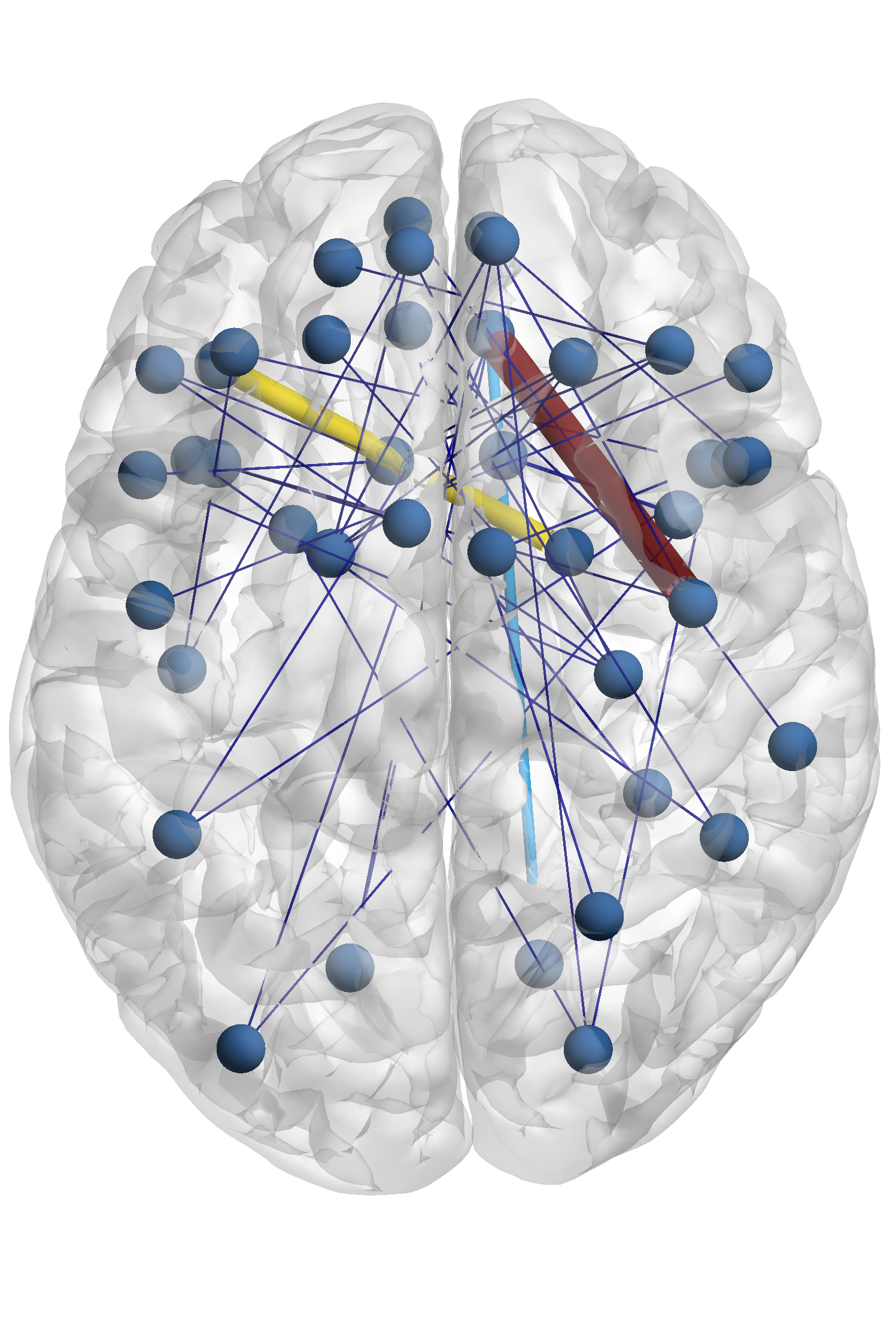}
	\includegraphics[scale=0.252]{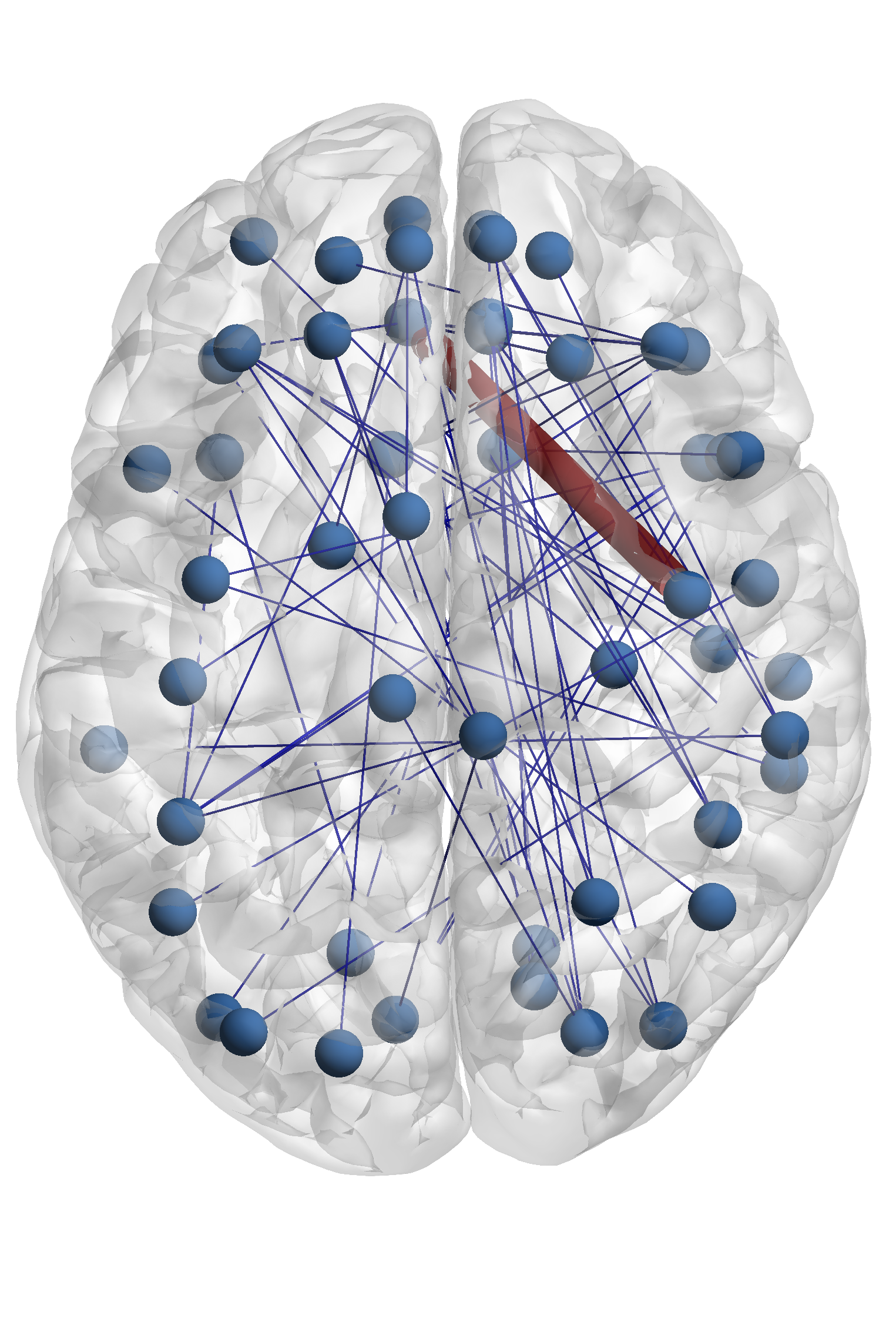}
	\includegraphics[scale=0.252]{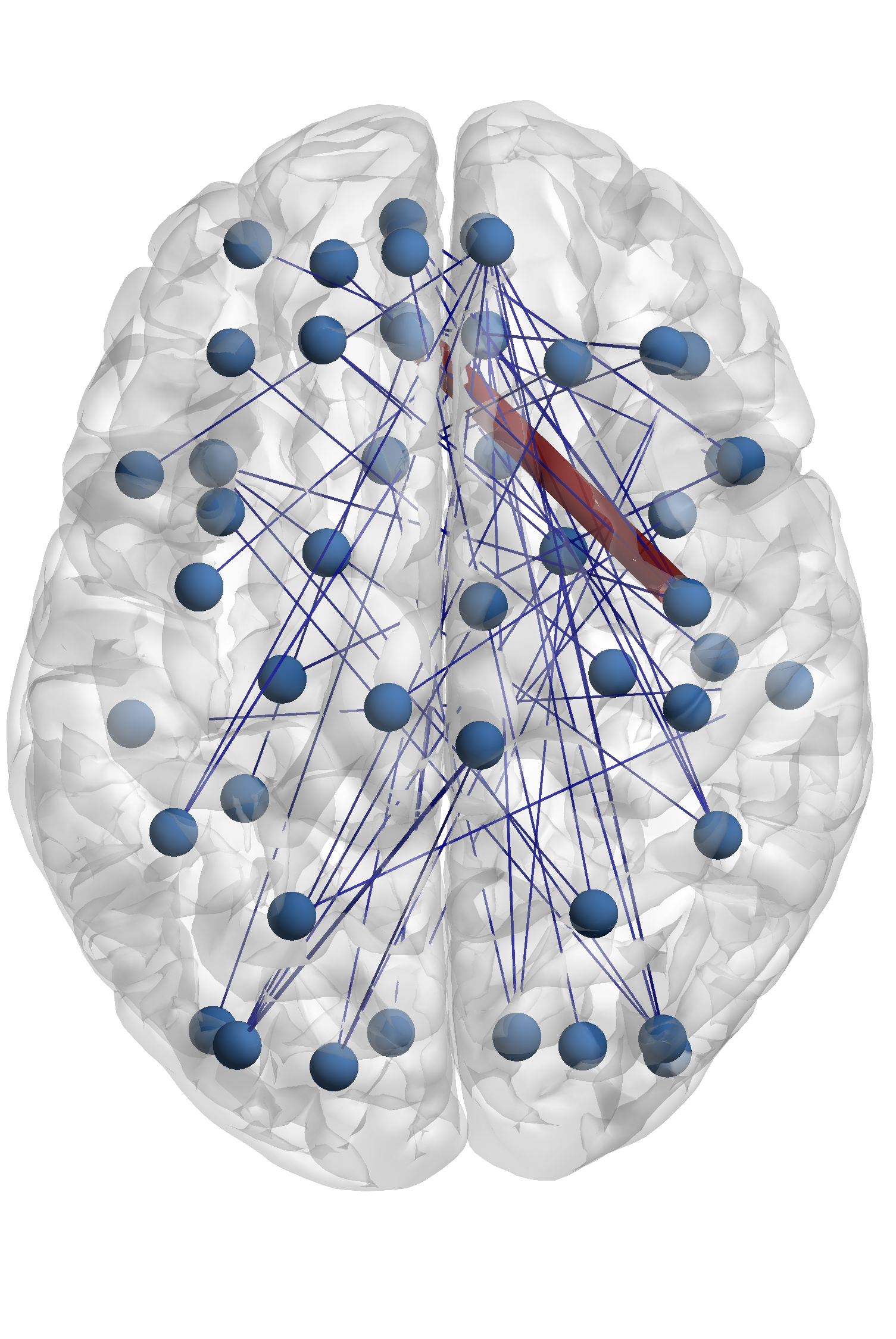}
	\includegraphics[scale=0.252]{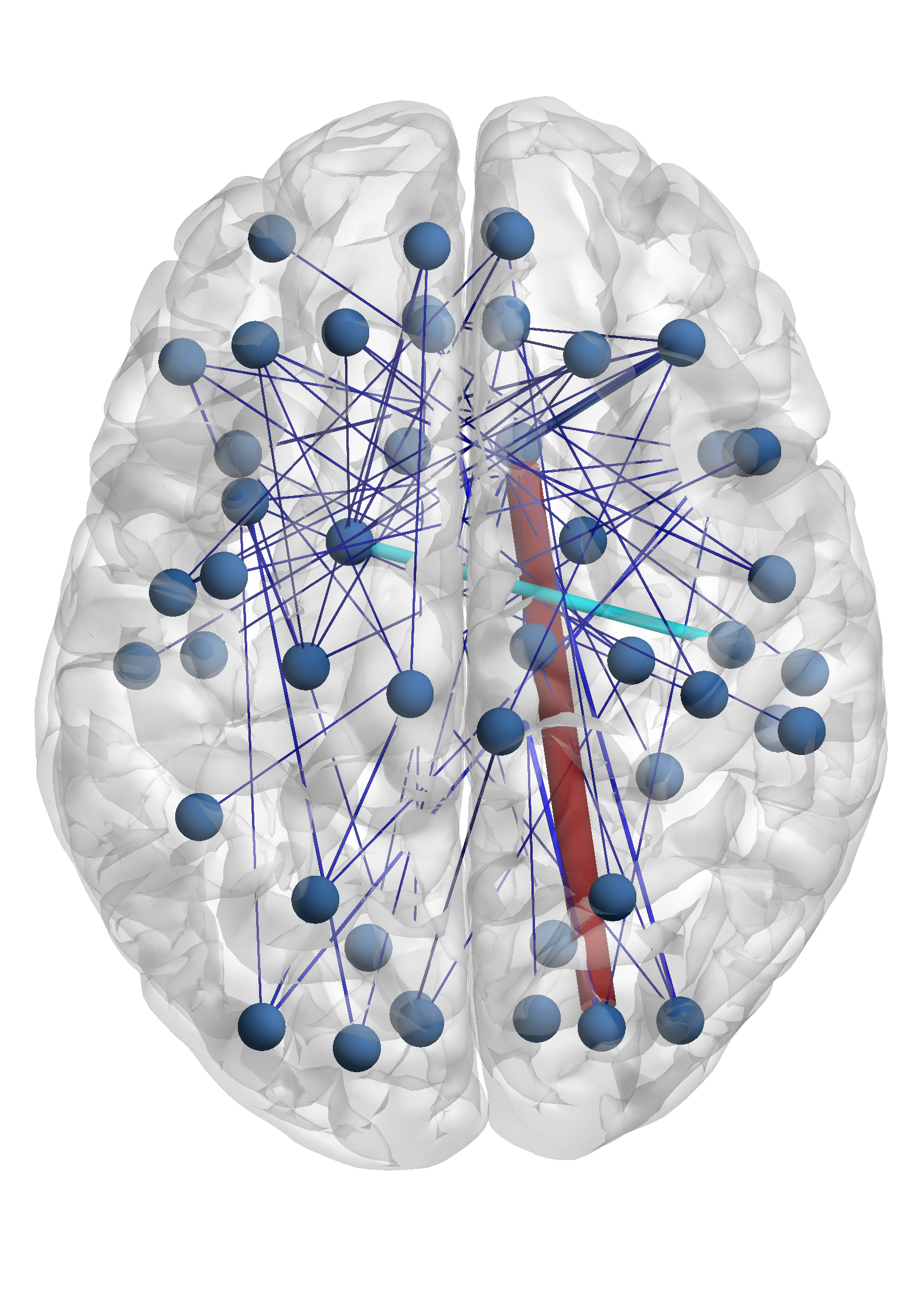}
	\includegraphics[scale=0.110]{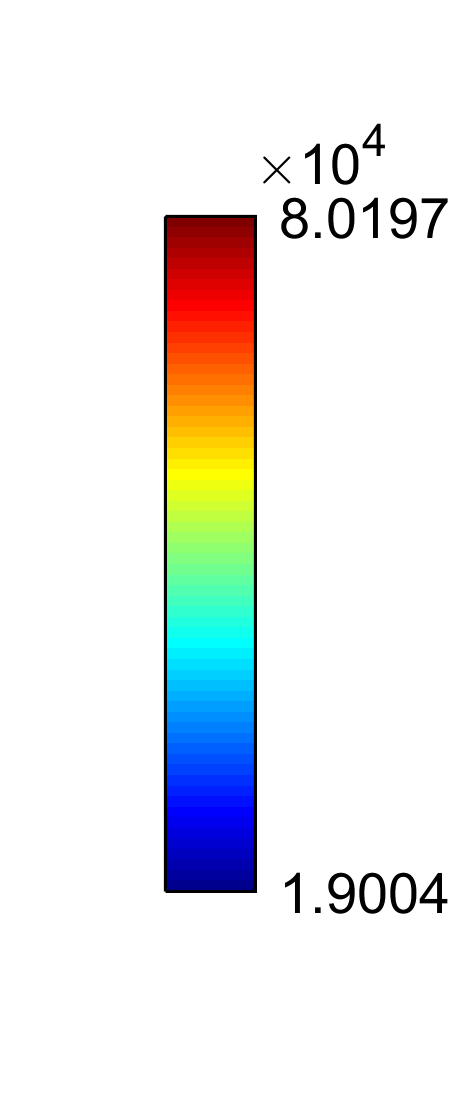}
	\caption{Precision of the Mesh Networks of a Subset of Subjects. The mesh networks from left to right indicate tasks in the following order: emotion, gambling, language, motor, relational, social and working-memory.}
	\label{fig:Precision}
\end{figure*}
\begin{table}
	{
		\centering
		\resizebox{1.\textwidth}{!}{
			\begin{minipage}{\textwidth}
				\begin{tabular}{ | c | c | c | c | c | c | c|}
		 \hline
		& Rand & A. Rand & & Rand & A. Rand \\
		\hline
		Raw fMRI Data &  0.68 & -0.07 &
		MAD   &  0.84  & 0.37  \\
		\hline	
				\end{tabular}
		\end{minipage}}
	}
	\caption{Clustering Performance Comparison.}
	\label{tab:table4}  
\end{table}
\begin{table}
{
	\centering
	\resizebox{1.2\textwidth}{!}{
 \begin{minipage}{\textwidth}
	\begin{tabular}{ | c | c | c | c | c | c | }
	\hline
	MAD & Rand & A. Rand & SDAE  & Rand & A. Rand\\
	\hline
	A0 &  0.84 & 0.37 &  A0 &  0.78 & 0.11\\
	\hline
	A1  &  0.83 & 0.34  &  A1  &  0.76  & 0.02\\
	\hline
	D1 &  0.81 & 0.28 &  D1 &  0.75 & -0.04\\
	\hline
	A2  &  0.77  & 0.15 &  A2  &  0.74  & -0.06\\
	\hline
	D2 &  0.86 & 0.47 &  D2 &  0.76 & 0.11\\
	\hline
	A3  &  0.75  & 0.12 &  A3  &  0.74  & 0.07\\
	\hline
	D3 &  0.72  & 0.15 &  D3 &  0.74 & -0.34\\
	\hline
	A4  &  0.68  & 0.06  &  A4  &  0.77  & 0.06\\
	\hline
	D4 &  0.77 & 0.24  &  D4 &  0.78 & 0.15\\
	\hline
	A5  &  0.68  & 0.08 &  A5  &  0.80  & 0.17\\
	\hline
	D5 &  0.74 & 0.17 &  D5 &  0.80 & 0.16\\
	\hline
	A6  &  0.75  & 0.18  &  A6  &  0.81  & 0.20\\
	\hline
	D6  &  0.75  & 0.17 &  D6  &  0.80  & 0.20\\
	\hline
	A7 &  0.87 & 0.50 &  A7 &  0.80 & 0.21\\
	\hline
	D7  &  0.84  & 0.37  &  D7  &  0.82  & 0.26\\
	\hline
	A8 &  0.85 & 0.37 &   A8 &  0.80 & 0.16\\
	\hline
	D8  &  0.82  & 0.27  &  D8  &  0.79  & 0.14\\
	\hline
	A9 & 0.85 & 0.39 &  A9 &  0.83 & 0.30\\
	\hline
	D9  &  0.82 & 0.28 &  D9  &  0.80  & 0.12\\
	\hline
	A10 &  0.82 & 0.29 &  A10 &  0.86 & 0.41\\
	\hline
	D10  &  0.83  & 0.30  &  D10  & 0.84  & 0.20\\
	\hline
	A11 &  0.79 & 0.20 &  A11 &  0.82 & 0.25\\
	\hline
	D11  &  0.81  & 0.26  &  D11  &  0.83  & 0.29\\
	\hline
	\end{tabular}
 \end{minipage}}
\caption{Clustering Performance for Individual Sub-bands.}
\label{tab:table2}  	
}\end{table}
\begin{table}
{
\centering
\resizebox{0.92\textwidth}{!}{
 \begin{minipage}{\textwidth}
	\begin{tabular}{ | l | c | c | l | c | c | }
	\hline
	MAD & Rand & A. Rand & SDAE  & Rand & A. Rand \\
	\hline
	All Sub-bands & 0.91 & 0.64 &  All Sub-bands&  0.93 & 0.71\\
	\hline
	Sub-bands 7-9  &  0.92 & 0.66  &  Subbands 7-9&  0.90 & 0.59 \\
	\hline
	Sub-bands 7-11 &  0.92 & 0.66 & Subbands 7-11& 0.91 & 0.60 \\
	\hline
	Sub-bands 3-8  &  0.89  & 0.57 & Subbands 3-8&  0.91  & 0.64 \\
	\hline
	Sub-bands 3-11 &  0.90 & 0.59 & Subbands 3-11&  0.91 & 0.63 \\
	\hline
	\end{tabular}
 \end{minipage}}
}
\caption{Clustering Performance for Combinations of Sub-bands.}
\label{tab:table3}  
\end{table}

Finally, we visualize the mesh networks obtained in the original fMRI signal to observe the inter-task and inter-subject variability of the brain networks. The motivation behind performing within-subject clustering rather than across-subject clustering in this study is the well-known inter-subject variability, which may prevent the clustering algorithm from finding natural groupings in the data. In order to illustrate the inter-subject variability, we plot the mesh networks of $3$ subjects in Fig. \ref{fig:2} and Fig \ref{fig:Precision} for each cognitive task. These subjects have the RI of $99\%$, which indicates that the proposed model has successfully estimated the natural groupings  for each one of these $3$ subjects. The networks shown in the aforementioned figures represent the medoids of the clusters which correspond to each one of the $7$ different tasks. The mesh networks corresponding to each of the subjects are pruned by eliminating the mesh arc weights with values less then a threshold to reach $1\%$ sparsity for simplification. A close analysis of the mesh networks corresponding to each task for the subjects shows that the mesh networks corresponding to the same task show small similarities across the $3$ subjects. This validates our prior claim on the existence of high inter-subject variabilities. To further investigate the inter-subject variability, we select a group of subjects with rand indices higher than $90\%$ from the HCP task data set of $300$ individuals. We define the precision of the mesh network across the set of subjects as the inverse of variance and calculate this value for the selected subjects. Fig. \ref{fig:Precision} shows the pruned precision of the mesh networks of the aforementioned set of subjects with $1\%$ sparsity. The thickness and the colors of the edges are proportional to their corresponding precision values. One may observe from Fig. \ref{fig:Precision} that the majority of the edges are thin-blue with only few of them thick-red. This indicates that the majority of the mesh network connections have high standard deviations across subjects.
\section{Conclusion} \label{sec:con}
In this paper, we proposed a framework for constructing a set of brain networks in multiple time-resolutions in order to model the connectivity patterns amongst the anatomic regions for different cognitive states. We proposed an unsupervised deep learning architecture that utilized these brain networks in multiple frequency sub-bands to learn the natural groupings of connectivity patterns in the human brain for given cognitive tasks. We showed that our suggested deep learning algorithm is capable of clustering the representative groupings into their associated cognitive states. We examined our suggested architecture on a task data set from HCP and achieved the clustering performance of $93\%$ Rand Index and $71\%$ Adjusted Rand Index for $200$ subjects. Lastly, we visualized the mean values and the precisions of the mesh networks at each component of the cluster mixture. We showed that the mean mesh networks at cluster centers have high inter-subject variabilities.
\section*{Acknowledgment}
We would like to thank Dr. Itir Onal Erturul, Arman Afrasiabi and Omer Ekmekci of Middle East Technical University and Professor Mete Ozay of Tohoku University for supporting us throughout many fruitful discussions. The work is supported by TUBITAK (Scientific and Technological Research Council of Turkey) under the grant No: 116E091.
\bibliographystyle{IEEEtran}
\bibliography{IEEEabrv,BibFile.bib}
\end{document}